%% file: main.tex
\documentclass{article}
\usepackage{ijcai19}

\usepackage{times}
\usepackage{soul}
\usepackage{url}
\usepackage[hidelinks]{hyperref}
\usepackage[utf8]{inputenc}
\usepackage[small]{caption}
\usepackage{graphicx}
\usepackage{amsmath}
\usepackage{booktabs}
\usepackage{algorithm}
\usepackage{algorithmic}
\usepackage{amsmath}
\usepackage{graphicx}

\usepackage{amsfonts}
\usepackage{amsmath}

\usepackage{bm}
\usepackage{bbm}

\usepackage{subfig}

\usepackage{multirow}
\usepackage{threeparttable}
\usepackage{tabularx}
\usepackage{paralist}

\usepackage{mleftright}

\usepackage{hyperref}       % hyperlinks
\usepackage{url}            % simple URL typesetting
\usepackage{booktabs}       % professional-quality tables
\usepackage{amsfonts}       % blackboard math symbols
\usepackage{nicefrac}       % compact symbols for 1/2, etc.
\usepackage{microtype}      % microtypography

\usepackage{wrapfig}

\usepackage{siunitx}
\usepackage{amsthm}
\usepackage[colorinlistoftodos]{todonotes}

\newcommand{\nop}[1]{}

\usepackage{xspace}
\newcommand{\model}{\textsc{UGraphEmb}\xspace}
\newcommand{\sdne}{\textsc{SDNE}\xspace}
\newcommand{\netmf}{\textsc{NetMF}\xspace}
\newcommand{\graphsage}{\textsc{GraphSAGE}\xspace}

\newcommand{\gk}{\textsc{Graph Kernels}\xspace}
\newcommand{\gv}{\textsc{Graph2Vec}\xspace}
\newcommand{\wv}{\textsc{Word2Vec}\xspace}
\newcommand{\beam}{\textsc{Beam}\xspace}
\newcommand{\hungarian}{\textsc{Hungarian}\xspace}
\newcommand{\vj}{\textsc{VJ}\xspace}
\newcommand{\hed}{\textsc{HED}\xspace}
\newcommand{\gcn}{\textsc{GCN}\xspace}
\newcommand{\gin}{\textsc{GIN}\xspace}

\newcommand{\ptc}{\textsc{Ptc}\xspace}
\newcommand{\imdb}{\textsc{ImdbM}\xspace}
\newcommand{\web}{\textsc{Web}\xspace}
\newcommand{\nci}{\textsc{Nci109}\xspace}
\newcommand{\reddit}{\textsc{Reddit12k}\xspace}

\DeclareMathOperator*{\concat}{\scalebox{1}[1.5]{$\parallel$}}
\newcommand\EE{\mathbb{E}}

\title{Unsupervised Inductive Graph-Level Representation Learning via \\ Graph-Graph Proximity}

\author{
Yunsheng Bai$^1$
\and
Hao Ding$^2$\and
Yang Qiao$^1$\and
Agustin Marinovic$^1$\and
Ken Gu$^1$\and \\
Ting Chen$^1$\and
Yizhou Sun$^{1}$\And
Wei Wang$^1$
\affiliations
$^1$University of California, Los Angeles\\
$^2$Purdue University
\emails
yba@ucla.edu,
ding209@purdue.edu,
angelinana0408@gmail.com,
amarinovic@ucla.edu,
ken.qgu@gmail.com,
\{tingchen, yzsun, weiwang\}@cs.ucla.edu
}

\begin{document}

\maketitle

\input{sec-abstract}
\input{sec-intro}

\input{sec-model}
\input{sec-exp}
\input{sec-related}

\input{sec-conc}

\subsubsection*{Acknowledgments}

This work is partially supported by NIH R01GM115833 and U01HG008488, NSF DBI-1565137, DGE-1829071, NSF III-1705169, NSF CAREER Award 1741634, and Amazon Research Award.

\clearpage
\small
\bibliographystyle{named}
\bibliography{bibliography}

\setcounter{section}{0} % restart numbering
\clearpage
\newpage
\section*{{\Large Supplementary Material}}

\def\thesection{\Alph{section}}

\input{sec-comp}

\input{sec-ged}
\input{sec-data}
\input{sec-data-preproc}
\input{sec-param-set}
\input{sec-analysis}

\input{sec-other-metric}

\input{sec-msna}

\input{sec-vis}

% \bibliographystyle{named}
% \bibliography{bibliography}

\end{document}

%% file: sec-abstract.tex
\begin{abstract}

We introduce a novel approach to graph-level representation learning, which is to embed an entire graph into a vector space where the embeddings of two graphs preserve their graph-graph proximity. Our approach, \model, is a general framework that provides a novel means to performing graph-level embedding in a completely unsupervised and inductive manner. The learned neural network can be considered as a function that receives any graph as input, either seen or unseen in the training set, and transforms it into an embedding. A novel graph-level embedding generation mechanism called Multi-Scale Node Attention (MSNA), is proposed. Experiments on five real graph datasets show that \model achieves competitive accuracy in the tasks of graph classification, similarity ranking, and graph visualization.
% \footnote{Anonymized link to supplementary material: \url{https://drive.google.com/drive/folders/1pVyPecm9A90x_mN1UAWq0TFyuF-IiuYH?usp=sharing}}.

% We also demonstrate that \model \yba{mention the pre-training and fine-tuning framework?}. We are among the first to evaluate unsupervised graph embeddings on the task of graph similarity ranking, and make the \yba{benchmark} datasets with ground-truth similarities publicly available\footnote{\url{https://drive.google.com/open?id=1pVyPecm9A90x_mN1UAWq0TFyuF-IiuYH}} for future researchers.

\end{abstract}

%% file: sec-intro.tex
\section{Introduction}
\label{sec-intro}

Recent years we have witnessed the great popularity of graph representation learning with success in not only node-level tasks such as node classification~\cite{kipf2016semi} and link prediction~\cite{zhang2018link}, but also graph-level tasks such as graph classification~\cite{ying2018hierarchical} and graph similarity/distance computation~\cite{bai2018graph}. 
% \YS{This sentence is too long, can you add punctuations?}

There has been a rich body of work~\cite{belkin2003laplacian,tang2015line,qiu2017network} 
% \WW{need to cite some node embedding papers here}
on node-level embeddings that turn each node in a graph into a vector preserving node-node proximity (similarity/distance). Most of these models are unsupervised and demonstrate superb performance in node classification and link prediction. It is natural to raise the question: Can we embed an entire graph into a vector in an unsupervised way, and how?
However, most existing methods for graph-level embeddings assume a supervised model~\cite{ying2018hierarchical,zhang2018capsule}, with only a few exceptions, such as \gk~\cite{yanardag2015deep} and \gv~\cite{narayanan2017graph2vec}. \gk typically count subgraphs for a given graph and can be slow. \gv is transductive (non-inductive), i.e. it does not naturally generalize to unseen graphs outside the training set. 
% \YS{Another long sentence. Can you please make them into different sentences?}
%Given the huge amount of works on unsupervised methods for node-level embeddings, i.e. turning each node into a vector using only the graph itself, it is natural to raise the question: How to embed an entire graph into a vector space in an unsupervised way?

% These vast amount of applications can be generally \yba{broadly / boiled down to / reduced to} divided into two categories: 
% \begin{enumerate}
% \item \textit{\textbf{Node-level tasks}}, such as node classification and link prediction. 
% \item \textit{\textbf{Graph-level tasks}}, such as graph classification and graph similarity/distance computation.
% \end{enumerate}

% For the graph-level tasks, especially classification, most existing works are supervised approaches, i.e. utilizing the graph labels to train a model, typically based on graph neural networks~\cite{ying2018hierarchical,zhao2018substructure,lee2018graph,zhang2018end}. However, the recent success of BERT~\cite{devlin2018bert}, an unsupervised pre-training method that achieves impressive state-of-the-art performance on many natural language tasks, \WW{inspired us to explore the potential of pre-training a neural network model for graph learning tasks.} 
% \yba{may be problematic since we are not that similar to BERT; need careful wording} 

A key challenge facing designing an unsupervised graph-level embedding model is the lack of graph-level signals in the training stage. \textit{Unlike node-level embedding which has a long history in utilizing the link structure of a graph to embed nodes, there lacks such natural proximity (similarity/distance) information between graphs}. Supervised methods, therefore, typically resort to graph labels as guidance and use aggregation based methods, e.g. average of node embeddings, to generate graph-level embeddings, with the implicit assumption that good node-level embeddings would automatically lead to good graph-level embeddings using only ``\textbf{\textit{intra}-graph} information'' such as node attributes, link structure, etc.

However, this assumption is problematic, as simple aggregation of node embeddings can only preserve limited graph-level properties, which is, however, often insufficient in measuring graph-graph proximity (``\textbf{\textit{inter}-graph}'' information).  
%when each graph is treated as a single data point and the relative position of multiple graphs is needed, such as in graph similarity search and graph alignment. 
Inspired by the recent progress on graph proximity modeling~\cite{ktena2017distance,bai2018graph}, we propose a novel framework, \emph{\textbf{\model}} ( \emph{\underline{U}}nsupervised \emph{\underline{Graph}}-level \emph{\underline{Emb}}bedding) that employs multi-scale aggregations of node-level embeddings, guided by the graph-graph proximity defined by well-accepted and domain-agnostic graph proximity metrics such as Graph Edit Distance (GED)~\cite{bunke1983distance}, Maximum Common Subgraph (MCS)~\cite{bunke1998graph}, etc.

% % \yba{mention the word ``graph-level''}

% In this paper, we make an initial step towards achieving good graph-level representations in an unsupervised manner, by only utilizing the graph-graph proximity. Such information can be obtained either by computing the proximity scores between a subset of graphs in a given dataset using general and well-defined metrics such as Graph Edit Distance (GED)~\cite{bunke1983distance}, Maximum Common Subgraph (MCS)~\cite{bunke1998graph}, etc. Alternatively, such information may come from some domain-specific metrics. For example, for detecting adverse drug-drug interaction~\cite{ma2018drug}, a domain-specific way to encode structure, called ECFP~\cite{rogers2010extended}, can be used to compute the proximity between drug molecules. In this work, we adopt the domain-agnostic metrics, GED and MCS, and leave the investigation of other graph proximity metrics as future work.

The goal of \model is to learn high-quality \textit{\textbf{graph-level}} representations in a completely \textit{\textbf{unsupervised}} and \textit{\textbf{inductive}} fashion: During training, it learns a function that maps a graph into a universal embedding space best preserving graph-graph proximity, so that after training, any new graph can be mapped to this embedding space by applying the learned function. 
Inspired by the recent success of pre-training methods in the text domain, such as \textsc{ELMO}~\cite{peters2018deep}, \textsc{Bert}~\cite{devlin2018bert}, and \textsc{GPT}~\cite{radford2018improving}. we 
    % make an initial step towards graph pre-training by
    further fine-tune the model via incorporating a supervised loss, to obtain better performance in downstream tasks, including but not limited to:

%The embeddings generated by \model can facilitate a series of downstream tasks:
\begin{itemize}
 \item \textbf{{Graph classification}}. The embeddings can be fed into any classification model such as logistic regression for graph classification.
    \item \textbf{{Graph similarity ranking}}. The embeddings learned by \model preserve the graph-graph proximity by design, and for a graph query, a ranked list of graphs that are similar to the query can be retrieved.
     
     \item \textbf{{Graph visualization}}. The embeddings can be projected into a 2-D space for graph visualization, where each graph is a point. It renders human insights into the dataset and facilitates further database analysis.

\end{itemize}

In summary, our contributions are three-fold:
\begin{enumerate}
    \item We formulate the problem of unsupervised inductive graph-level representation learning, and make an initial step towards pre-training methods for graph data. We believe that, given the growing amount of graph datasets of better quality, this work would benefit future works in pre-training methods for graphs.
    \item We provide a novel framework, \model, to generate \textbf{\textit{graph-level}} embeddings in a completely \textbf{\textit{unsupervised}} and \textbf{\textit{inductive}} fashion, well preserving graph proximity. A novel \textbf{Multi-Scale Node Attention (MSNA)} mechanism is proposed to generate graph-level embeddings. 
    \item We conduct extensive experiments on five real network datasets to demonstrate the superb quality of the embeddings by \model.
\end{enumerate}

% \yba{The rest of the paper is organized as follows...}

%% file: sec-model.tex
\section{The Proposed Approach: \model} %\YS

\begin{figure*}
\centering
\includegraphics[width=0.9\textwidth]{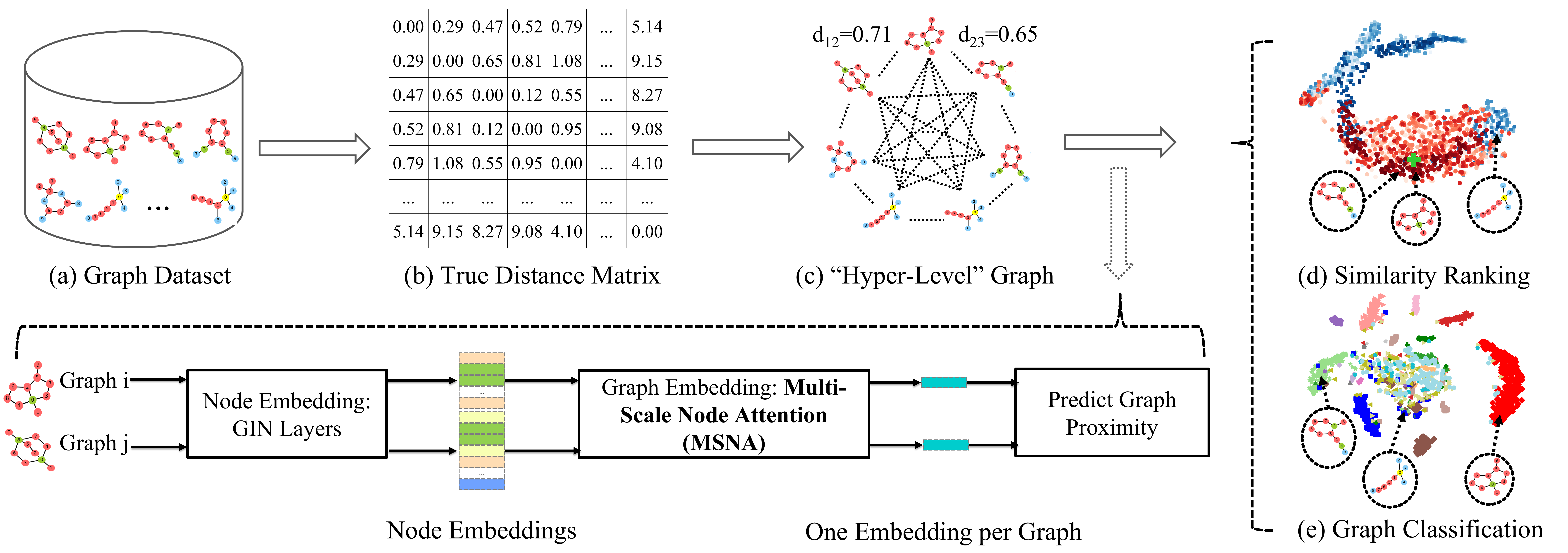}
\vspace*{-2mm}
\caption{Overview of \model. (a) Given a set of graphs, (b) \model first computes the graph-graph proximity scores (normalized distance scores in this example), (c) yielding a ``hyper-level graph'' where each node is a graph in the dataset, and each edge has a proximity score associated with it, representing its weight/strength. \model then trains a function that maps each graph into an embedding which preserves the proximity score. The bottom flow illustrates the details of graph-level embedding generation. (d) After embeddings are generated, similarity ranking can be performed. The green ``+'' sign denotes the embedding of an example query graph. Colors of dots indicate how similar a graph is to the query based on the ground truth (from red to blue, meaning from the most similar to the least similar). (e) Finally, \model can perform fine-tuning on the proximity-preserving graph-level embeddings, adjusting them for the task of graph classification. Different colors represent different graph labels in the classification task. 
% \YS{It seems to me (d) and (e) are two independent tasks. The current plot is misleading. A better way is to remove the arrow, and put them side-by-side. For example, we can move (d) and (e) to the bottom line, and move the architecture to the right hand side.} \yba{fixed}
% \WW{this caption is incomplete. We also need to add to the figure the node embeddings and multi-scale neighbor aggregations.} \yba{updated}
}
\label{fig:model}
\vspace*{-3mm}
\end{figure*}

We present the overall architecture of our unsupervised inductive graph-level embedding framework \model in Figure~\ref{fig:model}. The key novelty of \model is the use of graph-graph proximity. To preserve the proximity between two graphs, \model generates one embedding per graph from node embeddings using a novel mechanism called \textbf{Multi-Scale Node Attention (MSNA)}, and computes the proximity using the two graph-level embeddings. 
% The rest of the section describes these stages in detail \WW{this paragraph needs to be revised} \yba{updated}.

\subsection{Inductive Graph-Level Embedding}
% \YS{2.1 Inductive Graph-Level Embedding; 2.2 Unsupervised Loss via Inter-Graph Similarity Preservation}

\subsubsection{Node Embedding Generation}
% \YS{do not use ``stage"}

For each graph, \model first generates a set of node embeddings. There are two major properties that the node embedding model has to satisfy:
\begin{itemize}
    \item \textbf{{Inductivity}}.
    The model should learn a function such that for any new graph unseen in the training set, the learned function can be applied to the graph, yielding its node embeddings.
    \item \textbf{{Permutation-invariance}}.
    The same graph can be represented by different adjacency matrices by permuting the order of nodes, and the node embedding model should not be sensitive to such permutation.
\end{itemize}

Among various node embedding models, neighbor aggregation methods based on Graph Convolutional Networks (\gcn)~\cite{kipf2016semi} are permutation-invariant
% , i.e. robust to permutation of node ids,
and inductive.
% , i.e. can generalize to unseen nodes.
The reason is that the core operation, graph convolution, updates the representation of a node by aggregating the embedding of itself and the embeddings of its neighbors. Since the aggregation function treats the neighbors of a node as a set, the order does not affect the result. 

A series of neighbor aggregation methods have been proposed with different ways to aggregate neighbor information, e.g. \textsc{GraphSAGE}~\cite{hamilton2017inductive}, \textsc{GAT}~\cite{velickovic2017graph}, \gin~\cite{xu2018powerful}, etc. Since \model is a general framework for \textit{graph}-level embeddings, and all these models satisfy the two properties, any one of these methods can be integrated. We therefore adopt the most recent and state-of-the-art method, Graph Isomorphism Network (\gin)~\cite{xu2018powerful}, in our framework: 
% \yba{do we give an equation and/or insights for GIN or simply a citation? maybe refer to our sup material for details.}\WW{please include the equation and simple explanation} \yba{sure!}
\begin{align} \label{GIN-agg}
\bm{u_i}^{(k)} =   \mathrm{MLP}_{\bm{W_{k}}}^{(k)}   \left(  \left( 1 + \epsilon^{(k)} \right) \cdot \bm{u_i}^{(k-1)} +  \sum_{j \in \mathcal{N}(i)} \bm{u_j}^{(k-1)}\right)
\end{align}
where $\bm{u_i}$ is the representation of node $i$, $\mathcal{N}(i)$ is the set of neighbors of node $i$, $\mathrm{MLP}_{\bm{W_{k}}}^{(k)}$ denotes multi-layer perceptrons for the $k$-th \gin layer with learnable weights $\bm{W_{k}}$, and $\epsilon$ is a scalar that can either be learned by gradient descent or be a hyperparameter. \gin has been proven to be theoretically the most powerful GNN under the neighbor aggregation framework~\cite{xu2018powerful}.
% \yba{TODO: simple explanation}

\subsubsection{Graph Embedding Generation}

After node embeddings are generated, \model generates one embedding per graph using these node embeddings. Existing methods are typically based on aggregating node embeddings, by either a simple sum or average, or some more sophisticated way to aggregate. 
% The supplementary material provides a more detailed survey and explanation on some intuitions described next. 
% \YS{It seems we do not have supplementary material anymore in camera ready version?} \yba{commented}

However, since our goal is to embed each graph as a single point in the embedding space that preserves graph-graph proximity,
%where two graphs are closer if they are more similar under some graph proximity metric, 
the graph embedding generation model should:
\begin{itemize}
    \item \textbf{{Capture structural difference at multiple scales}}.
    Applying a neighbor aggregation layer on nodes such as \gin cause the information to flow from a node to its direct neighbors, so sequentially stacking $K$ layers would cause the final representation of a node to include information from its $K$-th order neighbors. 
    % In other words, the higher the number of neighbor aggregation layers, the larger the scale of the learned  embeddings.
    However, after many neighbor aggregation layers, the learned embeddings could be too coarse to capture the structural difference in small local regions between two similar graphs.
    %after aggregating neighbors multiple times, which is especially undesirable when the two graphs are very similar, and the structural difference lie in small local regions.
    Capturing structural difference at multiple scales is therefore important for \model to generate high-quality graph-level embeddings.
    
    \item \textbf{{Be adaptive to different graph proximity metrics}}.
    
    \model is a general framework that should be able to preserve the graph-graph proximity under any graph proximity metric, such as GED and MCS. A simple aggregation of node embeddings without any learnable parameters limits the expressive power of existing graph-level embedding models.
\end{itemize}

To tackle both challenges in the graph embedding generation layer, we propose the following \textbf{Multi-Scale Node Attention (MSNA)} mechanism. Denote the input node embeddings of graph $\mathcal{G}$ as $\bm{U_{\mathcal{G}}} \in \mathbb{R}^{N \times D}$, where the $n$-th row, $\bm{u_n} \in \mathbb{R}^{D}$ is the embedding of node $n$. The graph level embedding is obtained as follows:
% \WW{please explain Eq 2, especially where the novelty resides}
\begin{align} 
\label{eq:graph-level-emb}
\bm{h_{\mathcal{G}}} = \mathrm{MLP}_{\bm{W}} \left( \concat_{k=1}^K  {\mathrm{ATT}_{\bm{\Theta}^{(k)}}} (\bm{U_{\mathcal{G}}}) \right)
\end{align}
where $\parallel$ denotes concatenation, $K$ denotes the number of neighbor aggregation layers, $\mathrm{ATT}$ denotes the following multi-head attention mechanism that transforms node embeddings into a graph-level embedding, and $\mathrm{MLP}_{\bm{W}}$ denotes multi-layer perceptrons with learnable weights $\bm{W}$ applied on the concatenated attention results. 

The intuition behind Equation~\ref{eq:graph-level-emb} is that, instead of only using the node embeddings generated by the last neighbor aggregation layer, we use the node embeddings generated by each of the $K$ neighbor aggregation layers. 

$\mathrm{ATT}$ is defined as follows:
% \WW{does it have to be our previous model? You already adopt GIN in Stage 1. If the core of Stage 2 is also a previous model, the reviewer may feel this paper does not have enough novelty. We may want to think about how we present here.} \yba{sure. let me rephrase.}:
\begin{align}
\label{eq:att}
{\mathrm{ATT}_{\bm{\Theta}}} (\bm{U_{\mathcal{G}}}) 
= \sum_{n=1}^{N} \sigma(\bm{u_n}^{T} \mathrm{ReLU}(\bm{\Theta} (\frac{1}{N} \sum_{m=1}^{N} \bm{u_m}))) \bm{u_n}.
\end{align}
where $N$ is the number of nodes, $\sigma$ is the sigmoid function $\sigma(x)=\frac{1}{1+\exp{(-x)}}$, and $\bm{\Theta^{(k)}} \in \mathbb{R}^{D \times D}$ is the weight parameters for the $k$-th node embedding layer. 

The intuition behind Equation~\ref{eq:att} is that, during the generation of graph-level embeddings, the attention weight assigned to each node should be adaptive to the graph proximity metric. To achieve that, the weight is determined by both the node embedding $\bm{u_n}$, and a learnable graph representation.
% $\bm{c} \in \mathbb{R}^{D}$, 
% \WW{Why do we use $\bm{c}$ here, but not in Eq 3? We need to use consistent notations.} \yba{fixed}
The learnable graph representation is adaptive to a particular graph proximity via the learnable weight matrix $\bm{\Theta^{(k)}}$. 
% This provides the necessary learnability for \model to be adaptive to different graph proximity metrics. 
% A sigmoid function $\sigma(x)=\frac{1}{1+\exp{(-x)}}$ is applied to the result to ensure the attention weights is in the range $(0,1)$. We do not normalize the weights into length 1, since it is desirable to let the embedding norm reflect the graph size, which is essential for capturing the proximity between graphs. 

% Our supplementary material gives more discussion and some practical issues regarding these methods. \yba{TODO: mention the direction of supersource connection in the sup material.} 

\subsection{Unsupervised Loss via Inter-Graph Proximity Preservation}
% \YS{proximity: similarity or distance} \yba{should we replace similarity to distance everywhere? thank you very much!}\YS{proximity can be used to refer both similarity and distance. In the higher level, use proximity, when come to conrete measures, use similarity or distance.}

\subsubsection{Definition of Graph Proximity}
The key novelty of \model is the use of graph-graph proximity. It is important to select an appropriate graph proximity (similarity/distance) metric. We identify three categories of candidates:
\begin{itemize}
    \item \textbf{{Proximity defined by graph labels}}.
    
    For graphs that come with labels, we may treat graphs of the same label to be similar to each other. 
    However, such proximity metric may be too coarse, unable to distinguish between graphs of the same label. 
    %to for certain datasets graphs may have few to no labels, and what is worse, such proximity metric is too coarse -- graphs within the same class label cannot be further distinguished from each other.
    
    \item \textbf{{Proximity given by domain knowledge or human experts}}.
    
    For example, in drug-drug interaction detection~\cite{ma2018drug}, a domain-specific metric to encode compound chemical structure can be used to compute the similarities between chemical graphs. 
    However, such metrics do not generalize to graphs in other domains. Sometimes, this information may be very expensive to obtain. For example, to measure brain network similarities, a domain-specific preprocessing pipeline involving skull striping, band-pass filtering, etc. is needed. The final dataset only contains networks from 871 humans~\cite{ktena2017distance}.

    \item \textbf{{Proximity defined by domain-agnostic and well-accepted metrics}}.
    
    Metrics such as Graph Edit Distance (GED)~\cite{bunke1983distance} and Maximum Common Subgraph (MCS)~\cite{bunke1998graph} have been widely adopted in graph database search~\cite{yan2005substructure,liang2017similarity}, are well-defined and general to any domain. 
\end{itemize}

In this paper, we use GED as an example metric to demonstrate \model. 
GED measures the minimum number of edit operations to transform one graph to the other, where an edit operation on a graph is an insertion or deletion of a node/edge or relabelling of a node. Thus, the GED metric takes both the graph structure and the node labels/attributes into account. The supplementary material contain more details on GED. 
% \YS{(1) Again, supplementary material issue; (2) In our Arxiv paper, can we include additional experiments on MCS metric, which is already done in the rebuttal stage?} \yba{missing fine-tune exp; running now -- expected to be done by Friday morning}

\subsubsection{Prediction of Graph Proximity}
% \YS{It's distance not similarity} \yba{Do we change every similarity to distance in the entire paper? How to manage it?}

Once the proximity metric is defined, and the graph-level embeddings for $\mathcal{G}_i$ and $\mathcal{G}_j$ are obtained, denoted as $\bm{h}_{\mathcal{G}_i}$ and $\bm{h}_{\mathcal{G}_j}$, we can compute the similarity/distance between the two graphs.

\emph{Multidimensional scaling (MDS)} is a classic form of dimensionality reduction~\cite{williams2001connection}. The idea is to embed data points in a low dimensional space so that their pairwise distances are preserved, e.g. via minimizing the loss function
\begin{align} 
\mathcal{L}(\bm{h_i},\bm{h_j},d_{ij}) = (||\bm{h_i} - \bm{h_j}||_{2}^2 - d_{ij})^2
\end{align}
where $\bm{h_i}$ and $\bm{h_j}$ are the embeddings of points $i$ and $j$, and $d_{ij}$ is their distance. 

Since GED is a well-defined graph distance metric, we can minimize the difference between the predicted distance and the ground-truth distance:
\begin{align} 
\mathcal{L} &=
% \frac{1}{|\mathcal{D}|} \sum_
\EE_{(i, j) \sim \mathcal{D}} (\hat{d_{ij}} - d_{ij})^{2} \\
&= \EE_{(i, j) \sim \mathcal{D}} ( ||\bm{h_{\mathcal{G}_i}} - \bm{h_{\mathcal{G}_j}}||_{2}^2 - d_{ij})^{2}.
\end{align}
where $(i,j)$ is a graph pair sampled from the training set and $d_{ij}$ is the GED between them. 
% The supplementary materials describe the process of data preprocessing and sampling in detail.

Alternatively, if the metric is similarity, such as in the case of MCS, we can use the following loss function:
\begin{align} 
\mathcal{L} &= \EE_{(i, j) \sim \mathcal{D}} (\hat{s_{ij}} - s_{ij})^{2} \\
&= \EE_{(i, j) \sim \mathcal{D}} (\bm{h_{\mathcal{G}_i}}^{T} \bm{h_{\mathcal{G}_j}} - s_{ij})^{2}.
\end{align}

After training, the learned neural network can be applied to any graph, and the graph-level embeddings can facilitate a series of downstream tasks, and can be fine-tuned for specific tasks. For example, for graph classification, a supervised loss function can be used to further enhance the performance. 
% %while the model is still unsupervised.

%% file: sec-exp.tex
\section{Experiments} 
\label{sec-exp}

We evaluate our model, \model, against a number of state-of-the-art approaches designed for unsupervised node and graph embeddings, to answer the following questions:
\begin{enumerate}
\item[{\bf Q1}] How superb are the graph-level embeddings generated by \model, when evaluated with downstream tasks including graph classification and similarity ranking?
\item[{\bf Q2}] Do the graph-level embeddings generated by \model provide meaningful visualization for the graphs in a graph database? 
\item[{\bf Q3}] Is the quality of the embeddings generated by \model sensitive to choices of hyperparamters?
\end{enumerate}

% We provide an anonymized link to supplemental material containing details on datasets, baseline methods, and model configurations.

\textbf{Datasets} \enspace We evaluate the methods on five real graph datasets, \ptc, \imdb, \web, \nci, and \reddit.
% , whose detailed descriptions and statistics can be found in the supplementary material.
% \YS{supplementary material} \yba{fixed}
% \WW{we need to provide a table to show the statistics of the 5 datasets} \yba{nowadays a lot of papers put such table in the supplementary material, e.g. DiffPool, Capsule Graph Neural Networks}.
The largest dataset, \reddit, has 11929 graphs.

\subsection{Task 1: Graph Classification}

% To evaluate the embeddings generated by \model, we randomly split each dataset into training, validation, and testing sets by 6:2:2, followed by three stages: (1) train \model using the training set with validation set for parameter tuning; (2) train a standard logistic regression classifier using the embeddings as features as well as their ground-truth graph labels; (3) run \model on the graphs in the testing set and feed their embeddings into the classifier for label prediction. 

Intuitively, the higher the quality of the embeddings, the better the classification accuracy. Thus, we feed the graph-level embeddings generated by \model and the baselines into a logistic regression classifier to evaluate the quality: (1) \gk (\textsc{Graphlet} (\textsc{GK}), \textsc{Deep Graphlet} (\textsc{DGK}), \textsc{Shortest Path} (\textsc{SP}), \textsc{Deep Shortest Path} (\textsc{DSP}), \textsc{Weisfeiler-Lehman} (\textsc{WL}), and \textsc{Deep Weisfeiler-Lehman} (\textsc{DWL}))
% \YS{Full name and citation to these kernels, as pointed by the IJCAI reviewers.} \yba{fixed}
; (2) \gv~\cite{narayanan2017graph2vec}; (3) \netmf~\cite{qiu2017network}; (4) \graphsage~\cite{hamilton2017inductive}. 

For \gk, we also try using the kernel matrix and SVM classifier as it is the standard procedure outlined in \cite{yanardag2015deep}, and report the better accuracy of the two. For (3) and (4), we try different averaging schemes on node embeddings to obtain the graph-level embeddings and report their best accuracy.
% \WW{why do you list these 3 here? For node-embedding models, how do you compute the graph embedding? average?} \yba{see the commented section below. may not have space. plan is to put in sup material}.
% The supplementary material \YS{supplementary material} gives more details on the parameter settings.
% \yba{fixed}

% Since they only generate node-level embeddings, we take the average of node embeddings as the graph-level embedding. We also try various types of weighted averaging schemes, and report the best accuracy achieved by each of these baseline models.

% For the graph kernel baselines, there are two schemes to evaluate: (1) Treat the features extracted by each kernel method as the graph-level embeddings, and perform the second and third stages described previously; (2) Feed the SVM kernels generated by each method into a kernel SVM classifier as in \cite{yanardag2015deep}. The second scheme typically yields better accuracy and is more typical. We perform both schemes and report the better of the two accuracy scores for each baseline kernel.
% The key question is: \textit{Do good node-level embeddings automatically lead to good graph-level embeddings?}
As shown in Table~\ref{table:classif_results},
% \WW{this seems incomplete. We need more discussion of the results.} \yba{updated.}
% CapsGNN achieves top 2 on 6 out of 10 datasets and achieves comparable results on the other datasets.
% Compared with all the other end-to-end architectures, CapsGNN achieves top 1 on all the social datasets.
\model without fine-tuning, i.e. using only the unsupervised ``inter-graph'' information, can already achieve top 2 on 3 out of 5 datasets and demonstrates competitive accuracy on the other datasets. With fine-tuning (\textsc{\model-F}), our model can achieve the best result on 4 
% \YS{4} \yba{fixed}
out of 5 datasets. Methods specifically designed for graph-level embeddings (\gk, \gv, and \model) consistently outperform methods designed for node-level embeddings (\netmf and \graphsage), suggesting that \textit{good node-level embeddings do \textbf{not} naturally imply good graph-level representations}.

\begin{table}
\scriptsize
\centering
% \vspace{-0.05in}
  \begin{tabular}{cccccc}
    \toprule
    \textbf{Method} & \textbf{\ptc} & \textbf{\imdb} & \textbf{\web} & \textbf{\nci} & \textbf{\reddit} \\
      \midrule
      
    %  \multirow{4}{*}{\rotatebox{90}{\hspace*{-6pt}Kernel}} 
      
    \textsc{GK} & $57.26$ & $43.89$ & $21.37$ & $62.06$ & $31.82$ \\
    \textsc{DGK} & $57.32$ & $44.55$ & $23.65$ & $62.69$ & $32.22$ \\ 
    \textsc{SP} & $58.24$ & $37.01$ & $18.16$ & $73.00$ & $-$ \\
    \textsc{DSP} & $60.08$ & $39.67$ & $22.65$ & $73.26$ & $-$ \\
    \textsc{WL} & $66.97$ & $49.33$ & $26.44$ & $\textbf{80.22}$ & $39.03$ \\ 
    \textsc{DWL} & $70.17$ & $49.95$ & $34.56$ & $\textbf{80.32}$ & $39.21$ \\ \hline
    \gv & $60.17$ & $47.33$ & $\textbf{40.91}$ & $74.26$ & $35.24$ \\
    \netmf & $56.65$ & $30.67$ & $19.71$ & $51.84$ & $23.24$ \\
    \graphsage & $52.17$ & $34.67$ & $20.38$ & $65.09$ & $25.01$ \\
    % \arga & $-$ & $-$ & $-$ & $-$ & $-$ \\ 
    \hline
    % \textsc{\model-RANK} & $59.41$ & $\textbf{-}$ & $47.72$ & $\textbf{-}$ & $\textbf{-}$ & $\textbf{-}$ & $\textbf{-}$ & $\textbf{-}$ & $-$ & $-$ \\
    \textsc{\model} & $\textbf{72.54}$ & $\textbf{50.06}$ & $37.36$ & $69.17$ & $\textbf{39.97}$ \\
    \textsc{\model-F} & $\textbf{73.56}$ & $\textbf{50.97}$ & $\textbf{45.03}$ & $74.48$ & $\textbf{41.84}$ \\ \hline
    % \bottomrule
  \end{tabular}
%   \begin{tablenotes}
%   \end{tablenotes}
%   \vspace{-1.8em}
\centering
\caption{Graph classification accuracy in percent. ``-'' indicates that the computation did not finish after 72 hours. We highlight the top 2 accuracy in bold.}
\label{table:classif_results}
\end{table}

\begin{table*}
% \small
\centering
\vspace{-0.05in}
  \begin{tabular}{ccccccccccc}
    \toprule
    \multirow{3}{*}{\textbf{Method}} &
      \multicolumn{2}{c}{\textbf{\ptc}} &
      \multicolumn{2}{c}{\textbf{\imdb}} &
      \multicolumn{2}{c}{\textbf{\web}} &
      \multicolumn{2}{c}{\textbf{\nci}} &
      \multicolumn{2}{c}{\textbf{\reddit}} \\
      & \textbf{$\tau$} & \textbf{p@10} & \textbf{$\tau$} & \textbf{p@10} & \textbf{$\tau$} & \textbf{p@10} & \textbf{$\tau$} & \textbf{p@10} & \textbf{$\tau$} & \textbf{p@10} \\
      \midrule

    %  \multirow{4}{*}{\rotatebox{90}{\hspace*{-6pt}Kernel}} 
      
    \textsc{GK} & $0.291$ & $0.135$ & $0.329$ & $0.421$ & $0.147$ & $0.101$ & $0.445$ & $0.012$ & $0.007$ & $0.009$ \\
    \textsc{DGK} & $0.222$ & $0.103$ & $0.304$ & $0.410$ & $0.126$ & $0.009$ & $0.441$ & $0.010$ & $0.011$ & $0.012$ \\ 
    \textsc{SP} & $0.335$ & $0.129$ & $0.009$ & $0.011$ & $0.008$ & $0.065$ & $0.238$ & $0.012$ & $-$ & $-$ \\
    \textsc{DSP} & $0.344$ & $0.130$ & $0.007$ & $0.010$ & $0.011$ & $0.072$ & $0.256$ & $0.019$ & $-$ & $-$ \\
    \textsc{WL} & $0.129$ & $0.074$ & $0.034$ & $0.038$ & $0.014$ & $0.246$ & $0.042$ & $0.006$ & $0.089$ & $0.017$ \\ 
    \textsc{DWL} & $0.131$ & $0.072$ & $0.039$ & $0.041$ & $0.017$ & $0.262$ & $0.049$ & $0.009$ & $0.095$ & $0.023$ \\ \hline
    \gv & $0.128$ & $0.188$ & $0.697$ & $0.624$ & $0.014$ & $0.068$ & $0.033$ & $0.127$ & $0.008$ & $0.017$ \\
    \netmf & $0.004$ & $0.012$ & $0.003$ & $0.143$ & $0.002$ & $0.010$ & $0.001$ & $0.008$ & $0.009$ & $0.042$ \\
    \graphsage & $0.011$ & $0.033$ & $0.042$ & $0.059$ & $0.009$ & $0.010$ & $0.018$ & $0.040$ & $0.089$ & $0.017$ \\
    % \arga & $-$ & $-$ & $-$ & $-$ & $-$ & $-$ & $-$ & $-$ & $-$ & $-$ \\ 
    \hline
    \beam & $0.992^{*}$ & $0.983^{*}$ & $0.892^{*}$ & $0.968^{*}$ & $0.963^{*}$ & $0.957^{*}$ & $0.615^{*}$ & $0.997^{*}$ & $0.657^{*}$ & $1.000^{*}$ \\
    \hungarian & $0.755^{*}$ & $0.465^{*}$ & $0.872^{*}$ & $0.825^{*}$ & $0.706^{*}$ & $0.160^{*}$ & $0.667^{*}$ & $0.164^{*}$ & $0.512^{*}$ & $0.808^{*}$ \\
    \vj & $0.749^{*}$ & $0.403^{*}$ & $0.874^{*}$ & $0.815^{*}$ & $0.704^{*}$ & $0.151^{*}$ & $0.673^{*}$ & $0.097^{*}$ & $0.502^{*}$ & $0.867^{*}$ \\
    \hed & $0.788$ & $0.433$ & $0.627$ & $0.801$ & $\textbf{0.667}$ & $0.291$ & $0.199$ & $0.174$ & $0.199$ & $0.237$ \\ \hline
    % \textsc{\model-RANK} & $0.719$ & $0.403$ & $0.815$ & $0.582$ & $\textbf{-}$ & $\textbf{-}$ & $\textbf{-}$ & $\textbf{-}$ & $-$ & $-$ \\
    \textsc{\model} & $\textbf{0.840}$ & $\textbf{0.457}$ & $\textbf{0.853}$ & $\textbf{0.816}$ & $0.618$ & $\textbf{0.303}$ & $\textbf{0.476}$ & $\textbf{0.189}$ & $\textbf{0.572}$ & $\textbf{0.365}$ \\ \hline
    % \bottomrule
  \end{tabular}
%   \begin{tablenotes}
%   \end{tablenotes}
%   \vspace{-1.8em}
\centering
\caption{Similarity ranking performance. \beam, \hungarian, and \vj are three approximate GED computation algorithms returning upper bounds of exact GEDs. We take the minimum GED computed by the three as ground-truth GEDs for training and evaluating all the methods on both Task 1 and 2. Their results are labeled with ``$*$''. \hed is another GED solver yielding lower bounds. ``-'' indicates that the computation did not finish after 72 hours.}
\label{table:sim_results}
\end{table*}

% The supplementary material gives a more thorough analysis and discussion of the results.

\subsection{Task 2: Similarity Ranking}

For each dataset, we split it into training, validation, and testing sets by 6:2:2, and report the averaged \textit{Mean Squared Error (mse)}, \textit{Kendall's Rank Correlation Coefficient ($\tau$)}~\cite{kendall1938new}, and \textit{Precision at 10 (p@10)} to test the ranking performance. 
% The supplemental material \YS{supplemental material} shows more details on data preprocessing, baseline solvers, evaluation strategies, metrics, and model configurations.
% \yba{fixed}

Table~\ref{table:sim_results} shows that \model achieves state-of-the-art ranking performance under all settings except one. This should not be a surprise, because only \model utilizes the ground-truth GED results collectively determined by \beam~\cite{neuhaus2006fast}, \hungarian~\cite{riesen2009approximate}, and \vj~\cite{fankhauser2011speeding}. \model even outperforms \hed~\cite{fischer2015approximation}, a state-of-the-art approximate GED computation algorithm, under most settings, further confirming its strong ability to generate proximity-preserving graph embeddings by learning from a specific graph proximity metric, which is GED in this case.
% \WW{We need discussion of the results.} \yba{baselines are running.}

\subsection{Task 3: Embedding Visualization}

Visualizing the embeddings on a two-dimensional space is a popular way to evaluate node embedding methods~\cite{tang2015line}. However, we are among the first to investigate the question: Are the graph-level embeddings generated by a model like \model provide meaningful visualization?

We feed the graph emebddings learned by all the methods into the visualization tool t-SNE~\cite{maaten2008visualizing}. The three deep graph kernels, i.e. \textsc{DGK}, \textsc{DSP}, and \textsc{WDL}, generate the same embeddings as the non-deep versions, but use additional techniques~\cite{yanardag2015deep} to modify the similarity kernel matrices, resulting in different classification and ranking performance.

\begin{figure*}[h]
\centering
\includegraphics[scale=0.285]{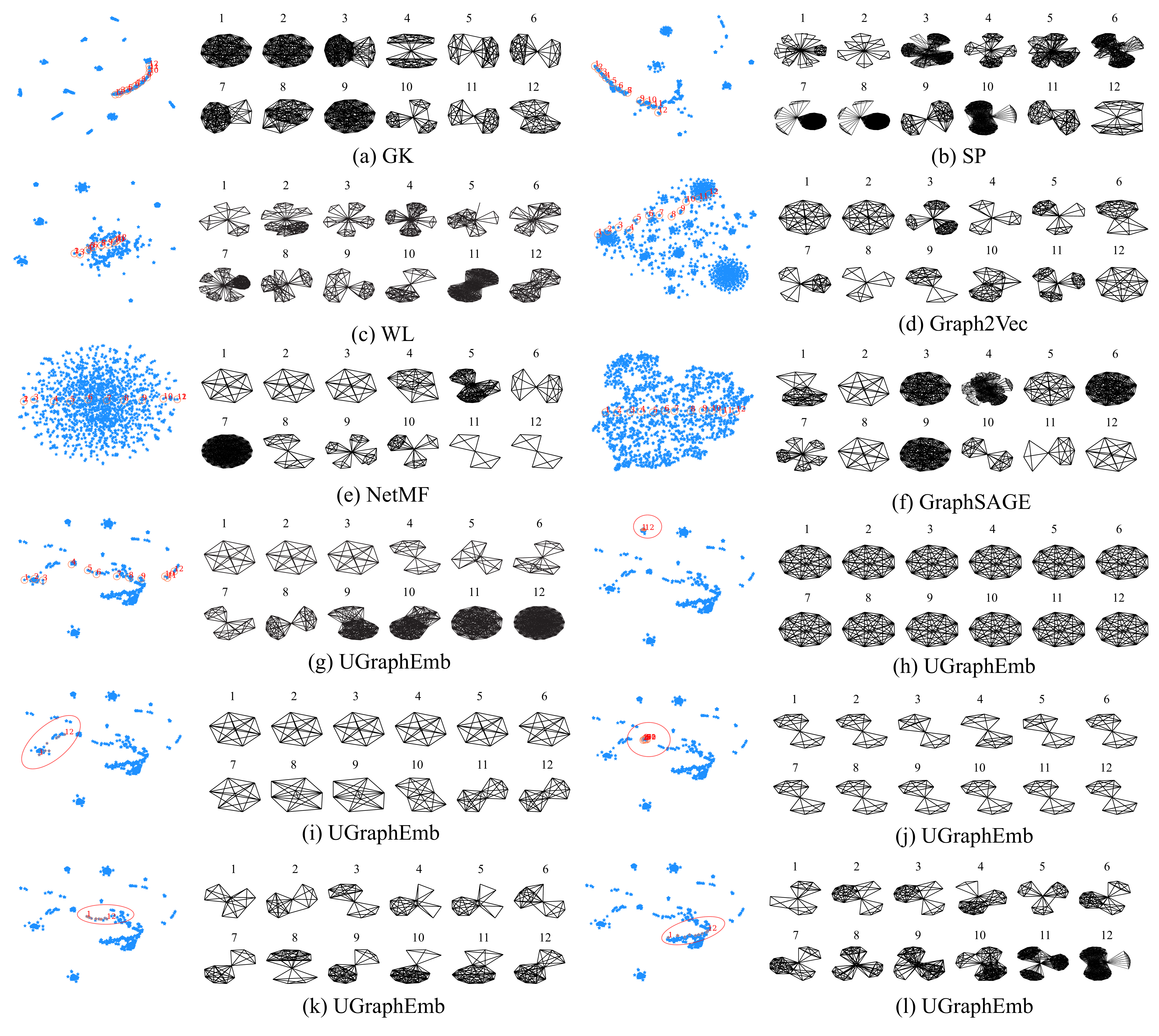}
% \caption{Visualization of the \imdb dataset. In the left column, for each method, 12 graphs are plotted. In the right column, we focus on \model: 5 clusters are highlighted in red circles. 12 graphs are sampled from each cluster and plotted to the right.}
\caption{Visualization of the \imdb dataset. From (a) to (g), for each method, 12 graphs are plotted. For (h) to (l), we focus on \model: 5 clusters are highlighted in red circles. 12 graphs are sampled from each cluster and plotted to the right.}
\label{fig:imdb_vis}
\vspace*{-3mm}
\end{figure*}

From Figure~\ref{fig:imdb_vis}, we can see that
% \WW{incomplete here} \yba{updated.}
\model can separate the graphs in \imdb into multiple clusters, where graphs in each cluster share some common substructures.

Such clustering effect is likely due to our use of graph-graph proximity scores, and is thus not observed in \netmf or \graphsage. For \gk and \gv though, there are indeed clustering effects, but by examining the actual graphs, we can see that graph-graph proximity is not well-preserved by their clusters (e.g. for \textsc{WL} graph 1, 2 and 9 should be close to each other; for \gv, graph 1, 2, and 12 should be close to each other), explaining their worse similarity ranking performance in Table~\ref{table:sim_results} compared to \model. 
% \YS{In order to best show this, we need to pick a local region, i.e., a cluster, to show that graphs inside the cluster are not similar to each other.} \yba{hard to find such. the real issue with WL and \gv is the similarity between clusters is not preserved. graphs in one cluster actually make sense.}

% \begin{figure*}
% \centering
% \includegraphics[scale=0.3]{ptc_vis}
% \caption{Visualization of the \ptc dataset.}
% \label{fig:ptc_vis}
% \vspace*{-2mm}
% \end{figure*}

% \begin{figure*}
% \centering
% \includegraphics[scale=0.3]{web_vis}
% \caption{Visualization of the \web dataset.}
% \label{fig:web_vis}
% \vspace*{-2mm}
% \end{figure*}

% \begin{figure*}
% \centering
% \includegraphics[scale=0.3]{nci_vis}
% \caption{Visualization of the \nci dataset.}
% \label{fig:nci_vis}
% \vspace*{-2mm}
% \end{figure*}

\subsection{Parameter Sensitivity of \model}

% \subsubsection{Parameter Sensitivity}
% \WW{This section seems from another paper.} \yba{Right. To be updated.} 
We evaluate how the dimension of the graph-level embeddings and the percentage of graph pairs with ground-truth GEDs used to train the model can affect the results. We report the graph classification accuracy on \imdb. 

\begin{figure}
   \centering
   {{\includegraphics[scale=0.26]{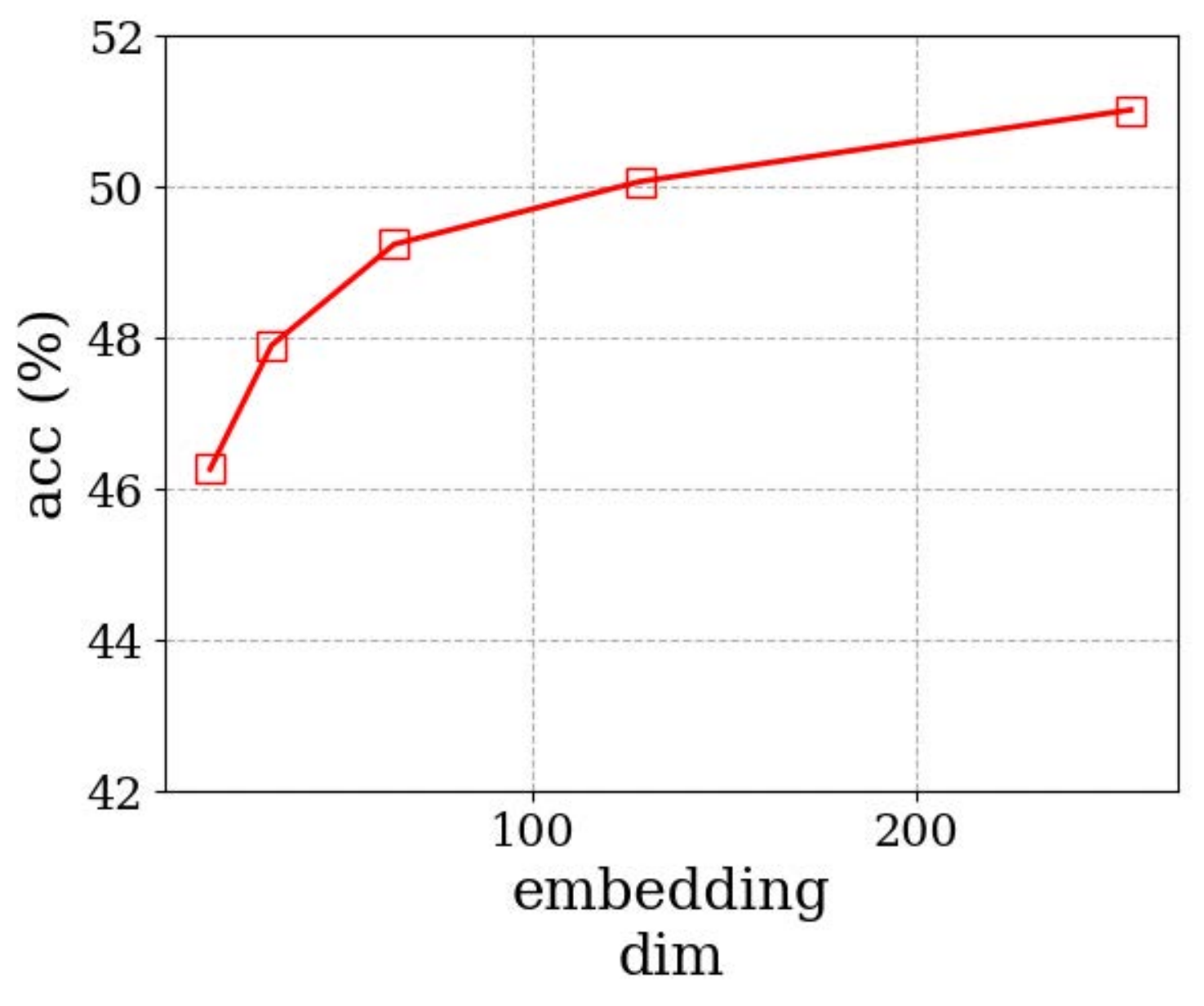}}}
   {{\includegraphics[scale=0.26]{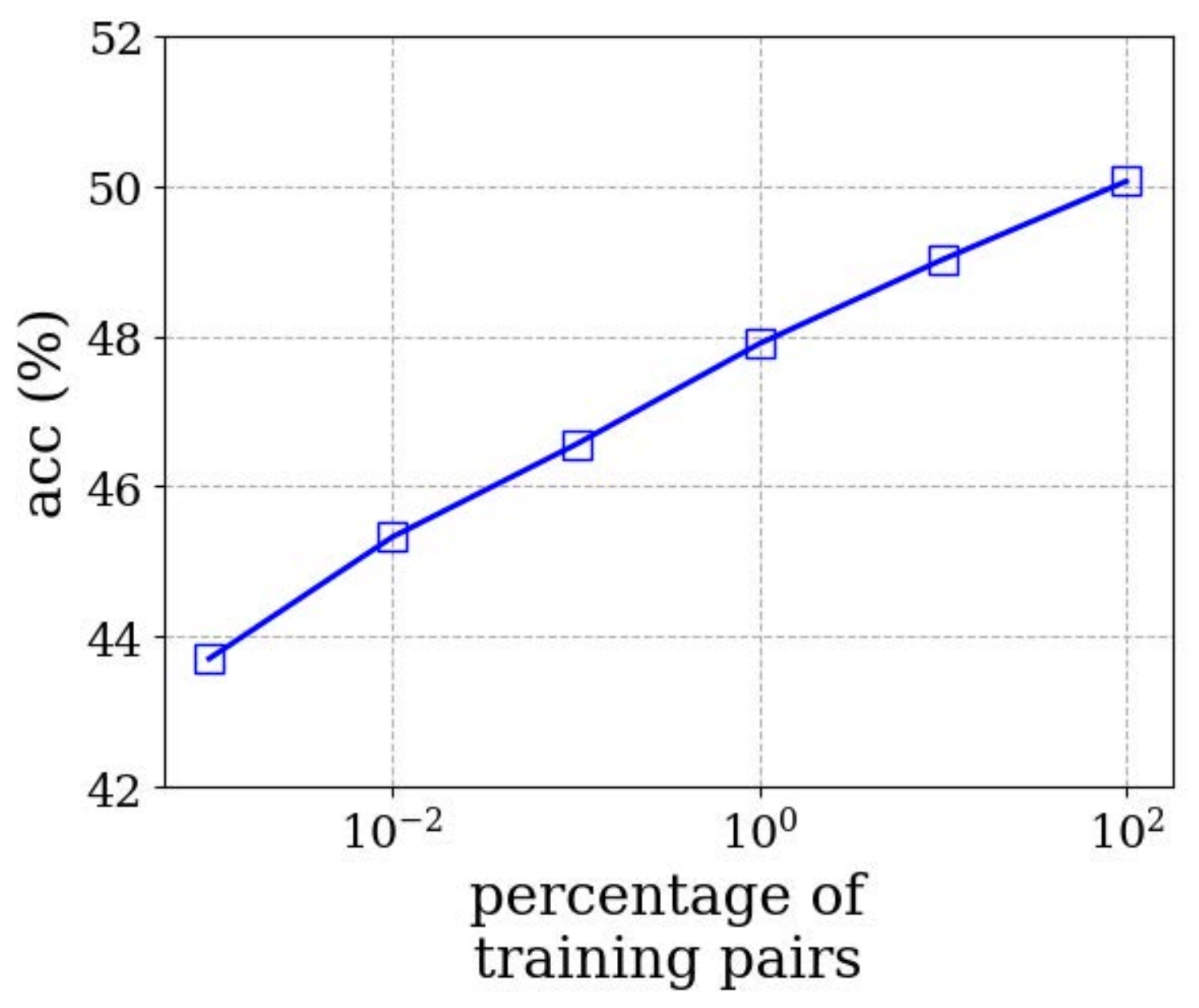}}}
   \caption{Classification accuracy on the \imdb dataset w.r.t. the dimension of graph-level embeddings and the percentage of graph pairs used for training.}
   \label{fig:param}
  \vspace*{-2mm}
\end{figure}

As can be seen in Figure~\ref{fig:param}, the performance becomes marginally better if larger dimensions are used. For the percentage of training pairs with ground-truth GEDs, the performance drops as less pairs are used. Note that the x-axis is in log-scale. When we only use 0.001\% of all the training graph pairs (only 8 pairs with ground-truth GEDs), the performance is still better than many baseline methods, exhibiting impressive insensitivity to data sparsity. 
% The supplementary material \YS{supplementary material. Please search the keywords to find others.} provides more details on training pair preparation.
% \yba{fixed}
% the rate of performance dropping is quite slow. 
% Even when only $0.001\%$ training pairs, i.e. $810000 * 0.00001 = 8$ pairs are used, 

%% file: sec-related.tex
\section{Related Work}
\label{sec-related}

% \subsection{Graph Representation Learning}

Unsupervised graph representation learning has a long history. Classic works including \netmf~\cite{qiu2017network}, LINE~\cite{tang2015line}, DeepWalk~\cite{perozzi2014deepwalk}, etc., which typically generate an embedding for each node in \textbf{\textit{one}} graph. Theoretical analysis shows that many of these works cannot handle embeddings for multiple graphs in the sense that the node embeddings in one graph are not comparable to those in another graph in any straightforward way~\cite{heimann2017generalizing}. A simple permutation of node indices could cause the node embedding to be very different.

More recently, some of the methods based on Graph Convolutional Networks (\gcn)~\cite{defferrard2016convolutional,kipf2016semi}, such as \textsc{VGAE}~\cite{kipf2016variational}, satisfy the desired permutation-invariance property.
% because \gcn essentially performs a local aggregation of neighbors for each node, so a simple permutation of the orders of a neighbor does not change its node embedding.
Categorized as ``graph autoencoders''~\cite{wu2019comprehensive}, they also belong to the family of graph neural network methods. Although satisfying the permutation-invariance requirement, these autoencoders are still designed to generate unsuperised node embeddings.

Methods designed for unsupervised graph-level embeddings include \gv~\cite{narayanan2017graph2vec}
% which treats each graph as a bag of rooted subgraphs and uses the \textsc{Doc2Vec} model~\cite{le2014distributed} to generate the embedding
and \gk~\cite{yanardag2015deep}, which however are either not based on learning or not inductive. 
% \YS{emphasize their limitation, not inductive, slow to compute?} \yba{updated}.
Unlike node-level information which is reflected in the neighborhood of a node,
% , i.e. the link structure of a graph, and the attributes associated with a node
graph-level information is much more limited. A large amount of graph neural network models resort to graph labels as a source of such information, 
%such information is ``intra-graph'', and 
making the models supervised aiming to improve graph classification accuracy specifically, such as \textsc{DiffPool}~\cite{ying2018hierarchical}, \textsc{CapsGNN}~\cite{zhang2018capsule}, etc.,
while \model learns a function that maps each graph into an embedding that can be used to facilitate many downstream tasks.

%% file: sec-conc.tex
\section{Conclusion}

% \WW{this need to be revised to echo back the key innovations we presented earlier in the paper.}
We present \model, an end-to-end neural network based framework aiming to embed an entire graph into an embedding preserving the proximity between graphs in the dataset under a graph proximity metric, such as Graph Edit Distance (GED). A novel mechanism for generating graph-level embeddings is proposed. Experiments show that the produced graph-level embeddings achieve competitive performance on three downstream tasks: graph classification, similarity ranking, and graph visualization. 

% In future, it would be interesting to explore possibilities of combining the unsupervised model with supervised loss, and \yba{TODO} to further enhance the performance of \model.

\label{sec-conc}

%% file: sec-comp.tex
\section{Comparison with Existing Frameworks}

\begin{figure*}
\centering
\includegraphics[scale=0.14]{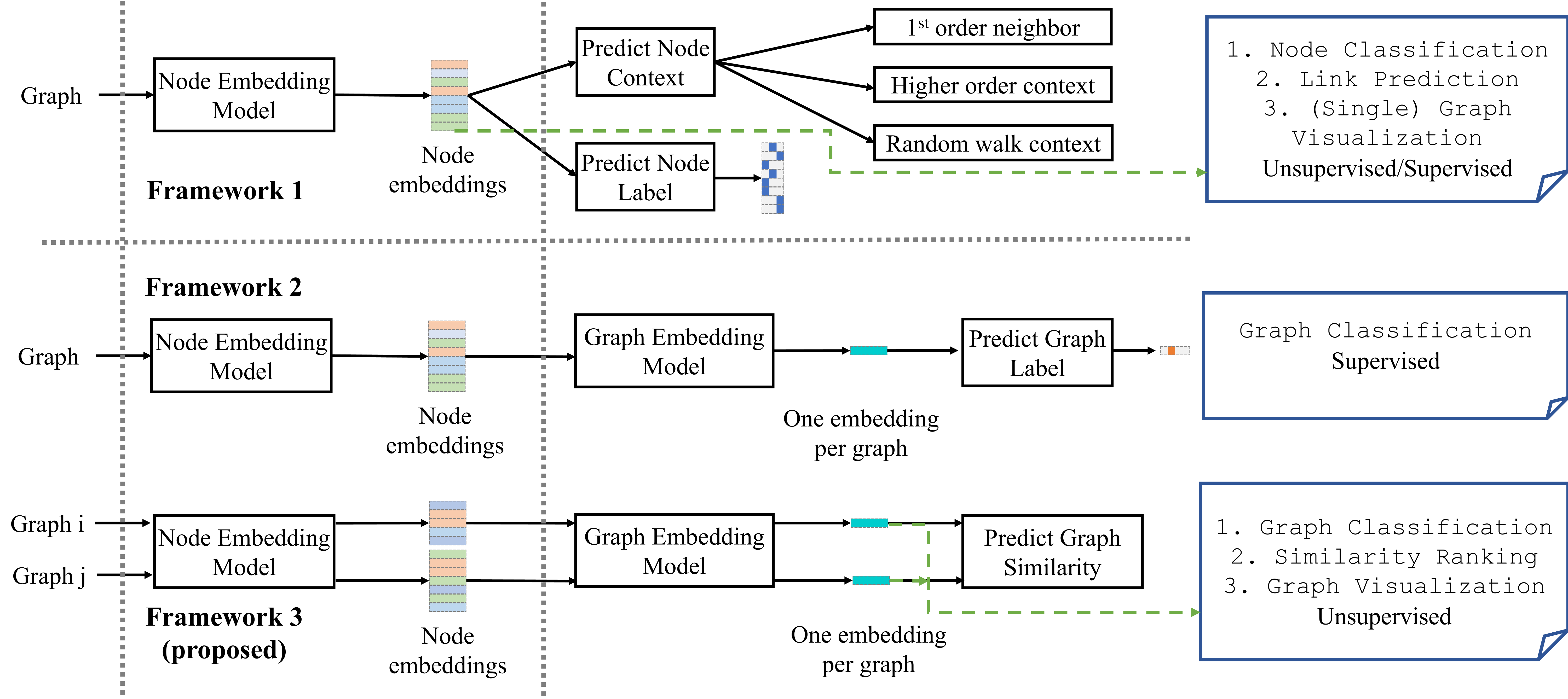}
\caption{Architecture 1 and 2 are typical neural network architectures for graph representation learning. Architecture 3 is our proposed \model.}
\label{fig:model}
\vspace*{-2mm}
\end{figure*}

To better see the novelty of our proposed framework, \model, we present a detailed study on two related existing frameworks for node and graph embeddings. As shown in Figure~\ref{fig:model}, we summarize graph neural network architectures for learning graph representations into three frameworks:
\begin{itemize}
    \item \textbf{{Framework 1}}: Supervised/Unsupervised framework for node-level tasks, e.g. node classification, link prediction, etc.
    \item \textbf{{Framework 2}}: Supervised end-to-end neural networks for graph-level tasks, typically graph classification.
    \item \textbf{{Framework 3}}: \model, unsupervised framework for multiple graph-level tasks with the key novelty in using graph-graph proximity.
\end{itemize}

The rest of this section describes the first two frameworks in detail, and compare \model with various other related methods when appropriate. This section also serves as a more thorough survey of graph embedding methods, proving more background knowledge in the area of node and graph representation learning. 

\subsection{Framework 1: Node Embedding (Supervised and Unsupervised)}
\label{subsec-framework-1}

Since the goal is to perform node-level tasks, the key is the ``Node Embedding Model'' which produces one embedding per node for the input graph. As described in the main paper, there are many methods to obtain such node embeddings, such as:
\begin{itemize}
    \item \textbf{{Matrix Factorization}}:
    
    This category includes a vast amount of both early and recent works on network (node) embedding, such as \textsc{Laplacian Eigenmaps (LLE)}~\cite{belkin2003laplacian}, \textsc{M-NMF}~\cite{wang2017community}, \textsc{NetMF}~\cite{qiu2017network}, etc. 
    
    Many interesting insights and theoretical analysis have been discovered and presented, but since this work focuses on neural network based methods, the reader is referred to \cite{qiu2017network} for a complete discussion.
    
    \item \textbf{{Direct Encoding (Free Encoder)}}: 
    
    This simple way of directly initializing one embedding per node randomly can be traced back to the Natural Language Processing domain -- the classic \wv~\cite{mikolov2013distributed} model indeed randomly initializes one embedding per word where gradients flow back during optimization. 
    
    Node embedding methods such as \textsc{LINE}~\cite{tang2015line} ans \textsc{DeepWalk}~\cite{perozzi2014deepwalk} adopt this scheme. \textsc{DeepWalk} is also known as ``skip-gram based methods''~\cite{du2018dynamic} due to its use of \wv.
    
    However, such methods are intrinsically transductive -- they cannot handle new node unseen in the training set in a straightforward way~\cite{hamilton2017inductive}. For \wv though, it is typically not a concern since out-of-vocabulary words tend to be rare in a large text corpus. 
    
    This also reveals a fundamental difference between the text domain and graph domain -- words have their specific semantic meaning making them identifiable across different documents, yet nodes in a graph or graphs usually lack such identity. This calls for the need of \textbf{{inductive representation learning}}, which is addressed more recently by the next type of node embedding model.
    
    \item \textbf{{Graph Convolution (Neighbor Aggregation)}}: 
    
    As discussed in the main paper, Graph Convolutional Network (\gcn)~\cite{defferrard2016convolutional} boils down to the aggregation operation that is applied to every node in a graph. This essentially allows the neural network model to \textbf{{learn a function that maps input graph to output node embeddings}}:
    \begin{align}
    \label{eq:inductive_func}
    \phi (\mathcal{G}) = \phi (\bm{A}_{\mathcal{G}}, \bm{F}_{\mathcal{G}}) = \bm{U}_{\mathcal{G}}.
    \end{align}
    where $\bm{A}_{\mathcal{G}}$ and $\bm{F}_{\mathcal{G}}$ denote the adjacency matrix (link structure) and the node and/or edge features/attributes, and $\bm{U_{\mathcal{G}}} \in \mathbb{R}^{N \times D}$ is the node embedding matrix. 
    
    The importance of such function $\phi$ is evident -- for any new node $i$ outside the training set of nodes, the neighbor aggregation models can simply apply the learned $\phi$ to obtain $\bm{u_i}$; for any new graph outside the training set of graphs, the same procedure also works, achieving \textbf{{inductivity}}. \textbf{{Permutation-invariance}} can also be achieved as discussed in the main paper.
    
    Methods including \textsc{GraphSAGE}~\cite{hamilton2017inductive}, \textsc{GAT}~\cite{velickovic2017graph}, \gin~\cite{xu2018powerful}, etc. are all under this category, with different aggregators proposed.
\end{itemize}

So far we have discussed about the node embedding generation step. In order to make the node embeddings high-quality, additional components are usually necessary as auxiliaries/guidance in the architectures of methods belong to Framework 1, including:
\begin{itemize}
    \item \textbf{{Predict Node Context}}:
    
    The goal is to use the node embeddings to reconstruct certain ``node local context'' -- in other words, to force the node embeddings to preserver certain local structure. We highlight three popular types of definitions of such context:
    \begin{itemize}
        \item \textbf{{1st order neighbor}}: 
        
        The model encourages directly connected nodes to have similar embeddings. In \textsc{LINE-1st}, the loss function is similar to the Skip-gram objective proposed in \wv. In \sdne~\cite{wang2016structural}, an auto-encoder framework, the loss function is phrased as the reconstruction loss, the typical name in auto-encoders. 
         
        \item \textbf{{Higher order context}}: 
        
        An example is \textsc{LINE-2nd}, which assumes that nodes sharing many connections to other nodes are similar to each other. In practice, such incorporation of higher order neighbors typically gives better performance in node-level tasks.
        
        \item \textbf{{Random walk context}}: 
        
        ``Context nodes'' are defined as the nodes that co-occur on random walks on a graph. By this definition, for a given node, its context can include both its close-by neighbors and distant node. 
        Equipped with various techniques of tuning and improving upon random walks, this type of methods seems promising.
        
        Example methods include \textsc{DeepWalk}, \textsc{Node2Vec}~\cite{grover2016node2vec}, \graphsage, etc. Notice that the former two use direct encoding as its node embedding model as described previously, while \graphsage uses neighbor aggregation. From this, we can also see that Framework 1 indeed includes a vast amount of models and architectures.
    \end{itemize}

% \begin{figure*}
% \centering
% \includegraphics[scale=0.26]{model_alter}
% \caption{Alternative view of \model. Given a set of graphs, \model first computes the graph-graph similarity scores, yielding a fully-connected hyper-level graph where each node is a graph in the dataset, and each edge has a similarity score associated with it representing its weight/strength. \model then trains a function that maps each graph into an embedding which preserves the similarity score. The green ``+'' sign denotes the embedding of an example query graph. Colors of dots indicate how similar a graph is to the query based on the ground truth (from red to blue, meaning from the most similar to the least similar). Finally, \model can perform fine-tuning on the similarity-preserving graph-level embeddings, adjusting them for the task of graph classification. Each color represents a graph label/class for the classification task. By fine-tuning for only a few \yba{TODO: check} iterations, state-of-the-art classification accuracy is achieved. Note: The graphs and plots are real data from our \web dataset as described in Section~\ref{sec-data}.}
% \label{fig:model_alter}
% \vspace*{-2mm}
% \end{figure*}

    \item \textbf{{Predict Node Label}}: 
    
    So far all the methods we have discussed about are unsupervised node embedding methods. As said in the main paper, to evaluate these unsupervised node embeddings, a second stage is needed, which can be viewed as a series of downstream  tasks as listed in Figure~\ref{fig:model}.
    
    However, a large amount of existing works incorporate a supervised loss function into the model, making the entire model trainable end-to-end. 
    
    Examples include \gcn as in \cite{kipf2016variational}, as well as a series of improvements upon \gcn, such as \textsc{GraphSAGE}~\cite{hamilton2017inductive}, \textsc{GAT}~\cite{velickovic2017graph}, \textsc{GIN}~\cite{xu2018powerful}, etc. as mentioned in the main paper.
\end{itemize}

Before finishing presenting Framework 1, we highlight important distinctions between the proposed framework and the following baseline methods:

\subsubsection{\model vs \netmf}

\netmf is among the state-of-the-art matrix factorization based methods for node embeddings. It performs eigen-decomposition, one-side bounding, and rank-$d$ approximation by Singular Value Decomposition, etc. for a graph, and is transductive. \model is graph-level and inductive. Section~\ref{subsec-baseline-setup} gives more details on how we obtain graph-level embeddings out of node-level embeddings for \netmf.

\subsubsection{\model vs \graphsage}

\graphsage belongs to neighbor aggregation based methods. 
Although being unsupervised and inductive, by design \graphsage performs node-level embeddings via an unsupervised loss based on context nodes on random walks (denoted as {\textbf{``Random walk context''}} as in Figure~\ref{fig:model}), while \model performs graph-level embeddings via the MSNA mechanism, capturing structural difference at multiple scales and adaptive to a given graph similarity/distance metric. 

% \subsubsection{\model vs \arga}

% \yba{TODO: we have to make sure we have \arga results from Brian first.}

\subsection{Framework 2: Supervised Graph Embedding}

The second framework we identify as supervised graph embedding. So far graph classification is the dominating and perhaps the only important task for Frmaework 2. 

Here we highlight some existing works to demonstrate its popularity, including \textsc{PatchySan}~\cite{niepert2016learning}, \textsc{ECC}~\cite{simonovsky2017dynamic}, \textsc{Set2Set}~\cite{gilmer2017neural}, \graphsage, \textsc{DGCNN/SortPool}~\cite{zhang2018end}, \textsc{SAN}~\cite{zhao2018substructure}, \textsc{DiffPool}~\cite{ying2018hierarchical}, \textsc{CapsGNN}~\cite{zhang2018capsule}, etc.

Notice that most of these models adopt the neighbor aggregation based node embedding methods described previously, which are inductive so that for new graphs outside the training set of graphs, their graph-level embeddings can be generated, ans graph classification can be performed.

\subsubsection{\model vs \gk}

Although \gk~\cite{yanardag2015deep} are not supervised methods, we still make a comparison here, because \gk are a family of methods designed for graph classification, the same task as Framework 2. 

Different graph kernels extract different types of substructures in a graph, e.g. graphlets~\cite{shervashidze2009efficient}, subtree patterns~\cite{shervashidze2009fast}, etc., and the resulting vector representation for each graph is typically called ``feature vector''~\cite{yanardag2015deep}, encoding the count/frequency of substructures. These feature vectors are analogous to graph-level embeddings, but the end goal is to create a kernel matrix encoding the similarity between all the graph pairs in the dataset fed into a kerkenl SVM classifier for graph classification. 

Compared with graph kernels, \model \textbf{{learns}} a function that preserves a general graph similarity/distance metric such as Graph Edit Distance (GED), and as a result, yields a graph-level embedding for each graph that can be used to facilitate a series of downstream tasks. It is inductive, i.e. handles unseen graphs due to the learned function. In contrast, although \gk can be considered as inductive~\cite{shervashidze2011weisfeiler}, graph kernels have to perform the subgraph extraction for every graph, which can be slow and cannot adapt to different graph proximity metrics.

\subsubsection{\model vs \gv}

Similar to \textsc{DeepWalk}, \gv is also inspired by the classic \wv paper, but instead of generating node embeddings, it is designed to generate graph-level embeddings, by treating each graph as a bag of rooted subgraphs, and adopting \textsc{Doc2Vec}~\cite{mikolov2013distributed} instead of \wv. The difference between \gv and \model is that, \gv is transductive (similar to \gk), as explained in Section~\ref{subsec-framework-1}, while \model is inductive. 

\subsection{Framework 3: \model}

This is our proposed framework, which is the key novelty of the paper. Now since we have introduced Framework 1 and Framework 2, it can be clearly seen that the use of graph-graph proximity is a very different and new perspective of performing graph-level embeddings. \model satifies all the following properties: {\textbf{graph-level}}, {\textbf{unsupervised}}, and {\textbf{inductive}}. Table~\ref{table:baseline_summary} shows a summary of the methods.
% Figure~\ref{fig:model_alter} illustrates Framework 3, \model, from the perspective of a graph dataset, instead of a graph pair as in Figure~\ref{fig:model}.

% The hyper-level graph in Figure~\ref{fig:model_alter} illustrates the analogy with Multidimensional Scaling \yba{as in the main paper (check! may move that section to here. we have potentially several things to say, and we need to organize which to put in main and which here)}. In future, it would be interesting to explore the possibilities of using strategies from Framework 1, especially those for the prediction of node local context, to further enhance the performance of \model.

% \subsubsection{\model vs \textsc{SimGNN}}

% \textsc{SimGNN}~\cite{bai2018graph} is a recent model designed for computing the pairwise graph similarity score. Compared to \textsc{SimGNN}, \model: (1) 

% \yba{concern: too similar to our wsdm paper; (1) goal; (2) multi-head sclae; (3) GIN instead of GCN; (4) BERT fine tune; (5) L-2 distance instead of NTN, instead of NTN, faster in real applications, because L-2 distance is faster and simpler, facilitating visualization (NTN not straightforward interpretability in the embedding space}

\begin{table}[h]
\small
% \scriptsize
\vspace*{-2mm}
% \scriptsize
\begin{tabular}
    {ccccc} \hline
    \textbf{Method} & \textbf{Citation} & \textbf{G} & \textbf{U} & \textbf{I} \\ \hline
    \textsc{LLE} & \cite{belkin2003laplacian} & $\times$ & $\checkmark$ & $\times$ \\ 
    % \textsc{PatchySan} & \cite{niepert2016learning} & $\checkmark$ & $\times$ & $\checkmark$ & $\times$ \\ 
    \gcn & \cite{kipf2016semi} & $\times$ & $\times$ & $\checkmark$ \\ 
    \gin & \cite{xu2018powerful} & $\times$ & $\times$ & $\checkmark$ \\ 
    \textsc{DiffPool} & \cite{ying2018hierarchical} & $\checkmark$ & $\times$ & $\checkmark$ \\ 
    % \textsc{CapsGNN} & \cite{zhang2018capsule} & $\checkmark$ & $\times$ & $\checkmark$ \\
    \hline
    \textsc{Graph Kernels} & - & $\checkmark$ & $\checkmark$ & $\checkmark$ \\
    \gv & \cite{narayanan2017graph2vec} & $\checkmark$ & $\checkmark$ & $\times$ \\
    % \sdne & \cite{wang2016structural} & $\times$ & $\checkmark$ & $\times$ & $\checkmark$ \\
    \netmf & \cite{qiu2017network} & $\times$ & $\checkmark$ & $\times$ \\ 
    \graphsage & \cite{hamilton2017inductive} & $\times$ & $\checkmark$ & $\checkmark$ \\
    % \arga & \cite{pan2018adversarially} & $\times$ & $\checkmark$ & $\checkmark$ & $\checkmark$ \\ 
    \hline
    \model & this paper & $$\checkmark$$ & $\checkmark$ & $\checkmark$ \\
    \hline
\end{tabular}
\centering
\caption{
% \WW{this table is not complete. We should at least include the ones we mentioned in the text, e.g., GCN, GIN} \yba{updated} 
A brief comparison of methods on node and graph representation learning. ``G'': Designed for graph-level embeddings ($\checkmark$) or not ($\times$). ``U'': Unsupervised or supervised. ``I'': Inductive or transductive. For \gk, there are multiple citations.
% \YS{why graph kernels not inductive?}
% \YS{B is not important at this stage and distracting.}
% \yba{
% From \cite{yang2016revisiting}, ``Since the embeddings are learned based on the graph structure, the above method is transductive, which means we can only predict instances that are already observed in the graph at training time. In many cases, however, it may be desirable to have an inductive approach, where predictions can be made on instances unobserved in the graph seen at training time''. 
% \gk extract subgraphs, which have to be done for test graphs and they have to compute the similarity between a test graph and all training graphs. Moreover, their embeddings are not even learned. We can use ``-'' to say that inductivity is not applicable for \gk.}
}
\label{table:baseline_summary}
\vspace*{-4mm}
\end{table}

\label{sec-related}

%% file: sec-ged.tex
\section{Graph Edit Distance (GED)}
\label{sec-func}

% \subsection{Definition of GED}

\begin{figure}[h]
\centering
\includegraphics[scale=0.17]{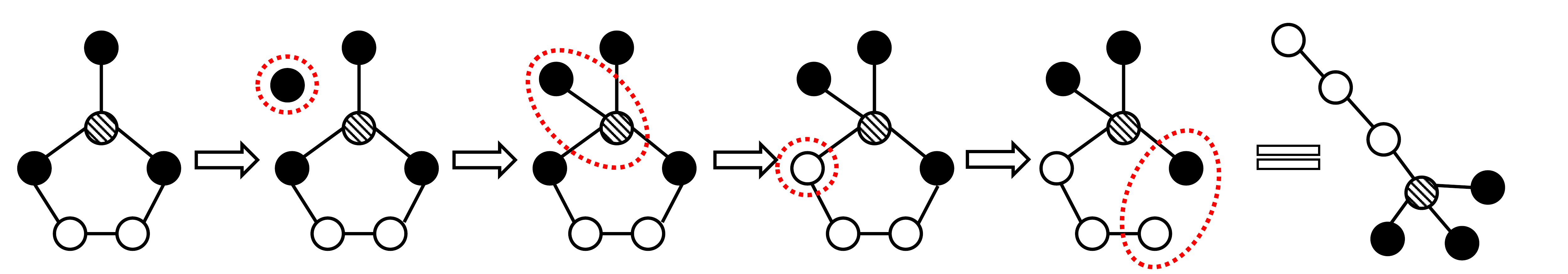}
\caption{The GED between the graph to the left and the graph to the right is 4, involving the following edit operation sequence: A node addition, an edge addition, a node label substitution, and an edge deletion.}
\label{fig:ged}
\vspace*{-2mm}
\end{figure}

The edit distance between two graphs~\cite{bunke1983distance} $\mathcal{G}_1$ and $\mathcal{G}_2$
% , denoted by $\mathrm{GED} (\mathcal{G}_1,\mathcal{G}_2)$,
is the number of edit operations in the optimal alignments that transform $\mathcal{G}_1$ into $\mathcal{G}_2$, where an edit operation on a graph $\mathcal{G}$ is an insertion or deletion of a node/edge or relabelling of a node. Note that other variants of GED definitions exist~\cite{riesen2013novel}, and we adopt the most basic version in this work.
% \footnote{Although other variants of GED exist~\cite{riesen2013novel}, we adopt this basic version.}.
Fig.~\ref{fig:ged} shows an example of GED between two simple graphs.

Notice that although \model currently does not handle edge types, \model is a general framework and can be extended to handle edge types, e.g. adapting the graph neural network described in \cite{kipf2018neural}.

%% file: sec-data.tex
\section{Datasets}
\label{sec-data}

\subsection{Detailed Description of Datasets}

\begin{table*}
% \vspace*{-4mm}
\begin{tabular}
{ccccccccc} \hline
\textbf{Dataset} & \textbf{Meaning} & \textbf{\#Node Labels} & \textbf{\#Graphs} & \textbf{\#Graph Labels} & \textbf{Min} & \textbf{Max} & \textbf{Mean} & \textbf{Std}\\ \hline
\textbf{\ptc} & Chemical Compounds & 19 & 344 & 2 & 2 & 109 & 25.5 & 16.2 \\
\textbf{\imdb} & Social Networks & 1 & 1500 & 3 & 7 & 89 & 13.0 & 8.5 \\
\textbf{\web} & Text Documents & 15507 & 2340 & 20 & 3 & 404 & 35.5 & 37.0 \\
\textbf{\nci} & Chemical Compounds & 38 & 4127 & 2 & 4 & 106 & 29.6 & 13.5 \\
\textbf{\reddit} & Social Networks & 1 & 11929 & 11 & 2 & 3782 & 391.4 & 428.7 \\
\hline
\end{tabular}
\centering
\caption{Statistics of datasets. ``Min'', ``Max'', ``Mean'', and ``Std'' refer to the minimum, maximum, mean, and standard deviation of the graph sizes (number of nodes), respectively.}
\label{table:dataset_summary}
% \vspace*{-2mm}
\end{table*}

Five real graph datasets are used for the experiments. A concise summary can be found in Table~\ref{table:dataset_summary}.

\begin{itemize}
     \item \ptc~\cite{shrivastava2014new} is a collection of 344 chemical compounds which report the carcinogenicity for rats. There are 19 node labels for each node.
    
    \item \imdb~\cite{yanardag2015deep} consists of 1500 ego-networks of movie actors/actresses with unlabeled nodes representing the people and edges representing the collaboration relationship. The nodes are unlabeled, but there could be 3 graph labels for each graph. 
    
    \item \web~\cite{riesen2008iam} is a collection of 2340 documents from 20 categories. Each node is a word, and there is an edge between two words if one word precedes the other. Since one word can appear in multiple sentences, the entire document is represented as a graph. Only the most frequent words are used to construct the graph, and there are 15507 words in total, thus 15507 node types associated with the dataset.
    
    \item \nci~\cite{wale2008comparison} is another bioinformatics dataset. It contains 4127 chemical compounds tested for their ability to suppress or inhibit human tumor cell growth.
    
    \item \reddit~\cite{yanardag2015deep} contains 11929 graphs each corresponding to an online discussion thread where nodes represent users, and an edge represents the fact that one of the two users responded to the comment of the other user. There is 1 of 11 graph labels associated with each of these 11929 discussion graphs, representing the category of the community.
\end{itemize}

\subsection{Additional Notes on \web}
\label{subsec-notes-web}

Since each graph node in \web represents a word, it is natural to consider incorporating the semantic similarity between two words, e.g. using \wv, into the GED definition, and even the broader topic of text matching and retrieval. 

In fact, the definition of GED does not specify that node labels must be discrete. There exists some variant of GED definition that can define node label difference in a more complicated way~\cite{riesen2013novel}, which is a promising direction to explore in future. It is also promising to explore document embedding based on graph representation of text.
% \yba{mention node features and GED and potential future work for text retreival and word-word similarities to be incoporated into GED def}

%% file: sec-data-preproc.tex
\section{Data Preprocessing}
\label{sec-data-preproc}

For each dataset, we randomly split 60\%, 20\%, and 20\% of all the graphs as training set, validation set, and testing set, respectively. For each graph in the testing set, we treat it as a query graph, and let the model compute the distance between the query graph and every graph in the training and validation sets. 

\subsection{Ground-Truth GED Computation}
\label{subsec-ged-algo}

To compute ground-truth GED for training pair generation as well as similarity ranking evaluation, we have the following candidate GED computation algorithms:
\begin{itemize}
    \item \textsc{A*}~\cite{hart1968formal}:
    
    It is an exact GED solver, but due to the NP-hard nature of GED, it runs in exponential time complexity. What is worse, a recent study shows that no currently available algorithm can reliably compute GED within reasonable time between graphs with more than 16 nodes~\cite{blumenthal2018exact}.
    
    \item \textsc{Beam}~\cite{neuhaus2006fast}, \textsc{Hungarian}~\cite{riesen2009approximate}, and \textsc{VJ}~\cite{fankhauser2011speeding}:
    
    They are approximate GED computation algorithms with sub-exponential time complexity, quadratic time complexity, and quadratic time complexity, respectively. They are all guaranteed to return upper bounds of the exact GEDs, i.e. their computed GEDs are always greater than or equal to the actual exact GEDs.
    
    \item \hed~\cite{fischer2015approximation}:
    
    It is another approximate GED solver running in quadratic time, but instead yields lower bounds of exact GEDs.
\end{itemize}

We take the minimum distance computed by \textsc{Beam}~\cite{neuhaus2006fast}, \textsc{Hungarian}~\cite{riesen2009approximate}, and \textsc{VJ}~\cite{fankhauser2011speeding}. The minimum is taken because their returned GEDs are guaranteed to be upper bounds of the true GEDs. In fact, the ICPR 2016 Graph Distance Contest~\footnote{\url{https://gdc2016.greyc.fr/}} also adopts this approach to handle large graphs.

We normalize the GEDs according to the following formula: $\mathrm{nGED}(\mathcal{G}_1,\mathcal{G}_2)=\frac{\mathrm{GED}(\mathcal{G}_1,\mathcal{G}_2)} {(|\mathcal{G}_1| + |\mathcal{G}_2|) / 2}$, where $|\mathcal{G}_i|$ denotes the number of nodes of $\mathcal{G}_i$~\cite{qureshi2007graph}.

% We use \textsc{Mcsplit}~\cite{mccreesh2017partitioning}, a recent exact MCS solver, to generate ground-truth for the MCS metric. Since MCS is already a graph similarity metric, we simply normalize the MCS size and treat the normalized MCS size as the true similarity score: $\mathrm{nMCS}(\mathcal{G}_1,\mathcal{G}_2)=\frac{|\mathrm{MCS}(\mathcal{G}_1,\mathcal{G}_2)|} {(|\mathcal{G}_1| + |\mathcal{G}_2|) / 2}$. 

% Please note that currently we are unaware of any approximate MCS algorithms that satisfy the exact MCS definition as used by the most state-of-the-art solver, \textsc{Mcsplit}. However, given the great efficiency improvement as shown in Section~\ref{subsec-efficiency} and the good accuracy as analyzed in Section~\ref{subsec-effectiveness}, \model demonstrates promising applicability in real world tasks where fast and accurate approximation is needed.

For the smaller datasets \ptc, \imdb, and \web, we compute the ground-truth GEDs for all the pairs in the training set. For the larger datasets \nci and \reddit, we do not compute all the pairs, and instead cap the computation at around 10 hours.

We run the ground-truth GED solvers on a CPU server with 32 cores, and utilize at most 20 cores at the same time, using code from \cite{riesen2013novel}. The details are shown in Table~\ref{table:training_pair_preparation}.

\begin{table}[H]
\small
% \vspace*{-4mm}
\begin{tabular}
{cccc} \hline
\textbf{Dataset} & \textbf{\#Total Pairs} & \textbf{\#Comp. Pairs} & \textbf{Time} \\ \hline
\textbf{\ptc} & 118336 & 42436 & 9.23 Mins \\
\textbf{\imdb} & 2250000 & 810000 & 4.72 Hours \\
\textbf{\web} & 5475600 & 1971216 & 8.23 Hours \\
\textbf{\nci} & 17032129 & 2084272 & 10.05 Hours \\
\textbf{\reddit} & 142301041 & 2124992 & 10.42 Hours \\
\hline
\end{tabular}
\centering
\caption{Number of graph pairs used to train \model on each dataset, along with the total wall time to compute the ground-truth GEDs for these pairs.}
\label{table:training_pair_preparation}
% \vspace*{-2mm}
\end{table}

\subsection{``Hyper-Level'' Graph}

% \yba{pairwise: training on small subset of pairs --> generalize/handle all graphs}
% \yba{mention fully-connected hyper-graph or not --> stochastic mini-batching training}
At this point, it is worth mentioning that the training procedure of \model is stochastic, i.e. \model is trained on a subset of graph pairs in each iteration. Moreover, \model does not require the computation all the graph pairs in the training set, so the notion of ``hyper-level'' graph as mentioned in the main paper does \textit{not} imply that \model constructs a fully connected graph where each node is a graph in the dataset.

In future, it would be promising to explore other techniques to construct such ``hyper-level'' graph beyond the current way of random selection of graph pairs in the training set.

\subsection{Node Label Encoding}
\label{subsec-nle}

% \yba{TODO: mention how to encode node labels initially}

For \ptc, \web, and \nci, the original node representations are one-hot encoded according to the node labels. For graphs with unlabeled nodes, i.e., \imdb and \reddit, we treat every node to have the same label, resulting in the same constant number as the initialize representation. 

In future, it would be interesting to consider more sophisticated ways to encode these node labels, because node labels help identifying different nodes across different graph datasets, which is an important component for a successful pre-training method for graphs. Consider that we want to combine multiple different graph datasets of different domains for large-scale pre-training of graph neural networks. Then how to handle different node labels in different datasets and domains becomes an important issue.

% \yba{mention why K-th order...}

% \yba{show red-blue plots}

% \yba{concern: too similar to our wsdm paper; (1) goal; (2) multi-head sclae; (3) GIN instead of GCN; (4) BERT fine tune; (5) L-2 distance instead of NTN, instead of NTN, faster in real applications, because L-2 distance is faster and simpler, facilitating visualization (NTN not straightforward interpretability in the embedding space}

%% file: sec-param-set.tex
\section{Parameter Settings and Experimental Details} 
\label{sec-param-set}

For the proposed model, to make a fair comparison with baselines, we use a single network architecture on all the datasets, and run the model using exactly the same test graphs as used in the baselines. 

We set the number of \gin layers to 3, and use ReLU as the activation function. The output dimensions for the 1st, 2nd, and 3rd layers of \gin are 256, 128, and 64, respectively. Following the original paper of \gin~\cite{xu2018powerful}, we fix $\epsilon$ to 0.

Then we transform the concreted embeddings into graph-level embeddings of dimension 256 by using two fully connected (dense) layers, which are denoted as $\mathrm{MLP}$ in the main paper. 

The model is written in TensorFlow~\cite{girija2016tensorflow}. We conduct all the experiments on a single machine with an Intel i7-6800K CPU and one Nvidia Titan GPU. As for training, we set the batch size to 256, i.e. 256 graph pairs (512 graphs) per mini-batch, use the Adam algorithm for optimization~\cite{kingma2014adam}, and fix the initial learning rate to 0.001. We set the number of iterations to 20000, and select the best model based on the lowest validation loss.

\subsection{Task 1: Graph Classification}

\subsubsection{Evaluation Procedure}

Since \model is unsupervised, we evaluate all the methods following the standard strategy for evluating unsupervised node embeddings~\cite{tang2015line,wang2016structural}. It has three stages: (1) Train a model using the training set with validation set for parameter tuning; (2) Train a standard logistic regression classifier using the embeddings as features as well as their ground-truth graph labels; (3) Run the model on the graphs in the testing set and feed their embeddings into the classifier for label prediction. 

\subsubsection{Baseline Setup}
\label{subsec-baseline-setup}

By default, we use the results reported in
the original work for baseline comparison. However, in cases where the results are not available, we use the code released by the original authors, performing a hyperparameter search based on the original author's guidelines. Notice that our baselines include a variety of methods of different flavors:

\begin{itemize}
    \item \gk:
    
    For the \gk baselines, there are two schemes to evaluate: (1) Treat the features extracted by each kernel method as the graph-level embeddings, and perform the second and third stages described previously; (2) Feed the SVM kernels generated by each method into a kernel SVM classifier as in \cite{yanardag2015deep}. The second scheme typically yields better accuracy and is more typical. We perform both schemes and report the better of the two accuracy scores for each baseline kernel. All the six versions of the \gk are described in detail in \cite{yanardag2015deep}.
    
    \item \gv: 
    
    Similar to \gk, \gv is also transductive, meaning it has to see all the graphs in both the training set and the testing set, and generates a graph-level embedding for each.
    
    \item \netmf and \graphsage:
    
    Since they generate node-level embeddings, we take the average of node embeddings as the graph-level embedding. We also try various types of averaging schemes, including weighted by the node degree, weighted by the inverse of node degree, as well as summation. We report the best accuracy achieved by these schemes.
    
    There is no training needed to be done for \netmf, since it is based on matrix factorization. For \graphsage, we combine all the graphs in the training set, resulting in one single graph to train \graphsage, which is consistent with its original design for inductive node-level embeddings. After training, we use the trained \graphsage model to generate graph-level embeddings for each individual graph in the test set, consistent with how \model handles graphs in the test set.
\end{itemize}

\subsubsection{Embedding Dimension}

For all the baseline methods, we ensure that the dimension of the graph-level embeddings is the same as our \model by setting hyperparameters for each properly. For \gk, however, they extract and count subgraphs, and for a given dataset, the number of unique subgraphs extracted depend on the dataset, which determines the dimension of the feature vector for each graph in the dataset. Thus, we do not limit the number of subgraphs they extract, giving them advantage, and follow the guidelines in their original papers for hyperparameter tuning. 

\subsubsection{Fine-Tuning}

To incorporate the supervised loss function (cross-entropy loss) into our model, we use multiple fully connected layers to reduce the dimension of graph-level embeddings to the number of graph labels. When the fine-tuning procedure starts, we switch to using the supervised loss function to train the model with the same learning rate and batch size as before. 

After fine-tuning, the graph label information is integrated into the graph-level embeddings. We still feed the embeddings into the logistic regression classifier for evaluation to ensure it is consistent for all the configurations of all the models. The accuracy based on the prediction of the model is typically much higher because it utilizes supervised information for graph label prediction. 

% Notice that such fine-tuning procedure could potentially be performed on the \graphsage baseline. However, \graphsage is originally designed for node-level embeddings. With an aggregation based framework \yba{TODO}  and \model already outperforms \graphsage under the unsupervised setting,  

\subsection{Task 2: Similarity Ranking}

\subsubsection{Evaluation Procedure}

For all the methods, we adopt the procedure outlined in Section~\ref{subsec-baseline-setup} to obtain graph-level embeddings. For each graph in the test set, we treat it as a graph query, compute the similarity/distance score between the query graph and every graph in the training set, and rank the results, compared with the ground-truth ranking results by the ground-truth GED solvers.

\begin{figure*}
\centering
\includegraphics[scale=0.8]{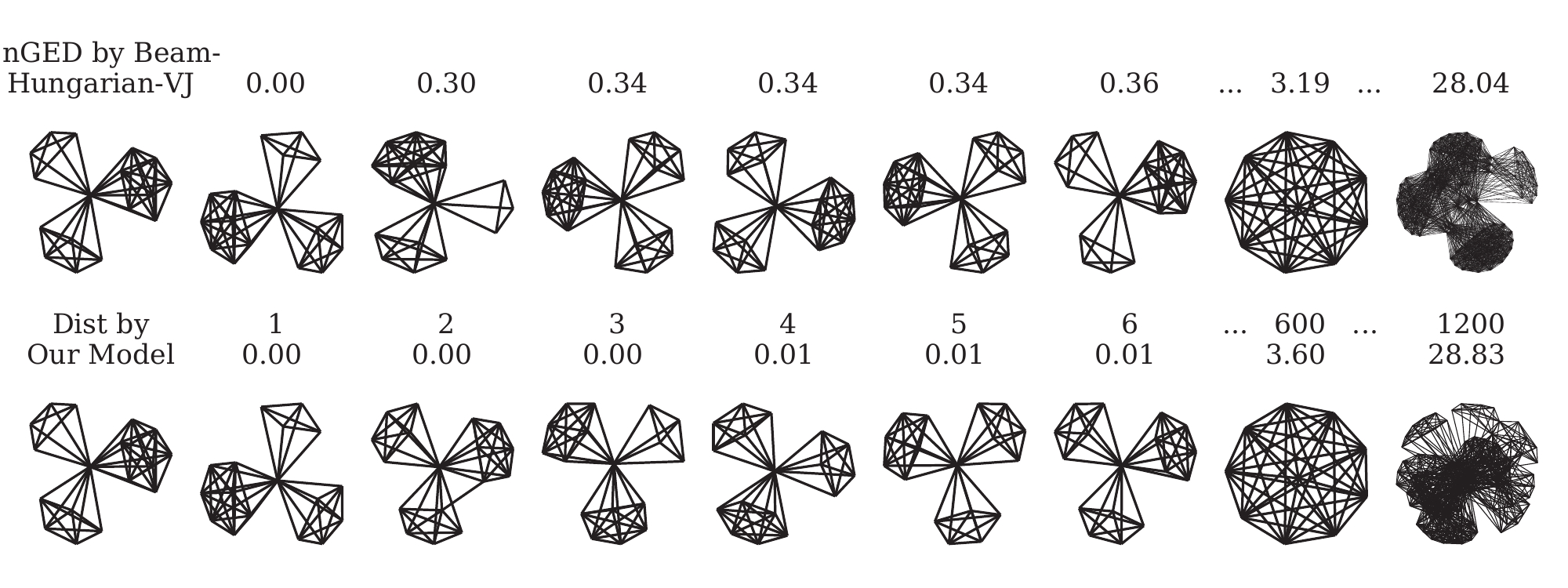}
\caption{Visualization of the ranking results on a query in \imdb. The top row depicts the ground-truth ranking, labeled with normalized ground-truth GEDs, and the bottom row depicts ranking by \model, depicts the distance score computed by \model.}
\label{fig:query_imdb_vis}
\vspace*{-2mm}
\end{figure*}

We compute both the similarity score (inner product of two graph-level embeddings) and distance score (squared L-2 distance between two graph-level embeddings) for every graph pair when doing the query, and report the better of the two in the paper. To verify that the actual ranking of the graphs makes sense, we perform several case studies. As shown in Figure~\ref{fig:query_imdb_vis}, \model computes the distance score between the query and every graph in the training set. Although the exact distance score is not exactly the same as the ground-truth normalized GEDs, the relatively position and ranking are quite reasonable.

Notice that for \gk, the three deep versions, i.e. \textsc{DGK}, \textsc{DSP}, and \textsc{WDL}, generate the same graph-level embeddings as the non-deep versions, i.e. \textsc{GK}, \textsc{SP}, and \textsc{DL}, but use the idea of \wv to model the relation between subgraphs~\cite{yanardag2015deep}. Consequently, the non-deep versions simply compute the dot products between embeddings to generate the kernel matrices, but the deep versions further modify the kernel matrices, resulting in different graph-graph similarity scores. We thus evaluate the deep versions using their modified kernel matrices.

\subsection{Task 3: Embedding Visualization}

\subsubsection{Evaluation Procedure}

As outlined in the main paper, we feed the graph-level embeddings into the t-SNE~\cite{maaten2008visualizing} tool to project them into a 2-D space. We then do the following linear interpolation: (1) Select two points in the 2-D space; (2) Form a line segment between the two selected points; (3) Split the line into 11 equal-length line segments, resulting in 12 points on the line segment in total; (4) Go through these 12 points: For each point, find an embedding point in the 2-D space that is closest to it; (5) Label the 12 embedding points on the embedding plot and draw the actual graph to the right of the embedding plot. This yields the visualization of the \imdb dataset in the main paper.

%% file: sec-analysis.tex
\section{Analysis and Discussion of Experimental Results}
\label{sec-analysis}

% \yba{address without fine-tuning, embedding may be bad for graph classification}

% \yba{explain why fine-tune does not achieve SOTA}

% \yba{tsne baselines have their own def of proximity, but is not... (1) not well accepted; (2) slow to compute: (3) not flexible/adaptive; (4) classification perf we are also good}

On graph classification, \model does not achieve top 2 on \web and \nci, which can be attributed to the fact that there are many node labels associated with the two datasets, as shown in Table~\ref{table:dataset_summary}. Combined with the fact that we use one-hot encoding for the initial node representations as described in Section~\ref{subsec-nle}, \model has limited capacity to capture the wide variety of node labels. In future, it is promising to explore other node encoding techniques, such as encoding based on node degrees and clustering coefficients~\cite{ying2018hierarchical}.

Another possible reason is that the current definition of GED cannot capture the subtle difference between different node labels. For example, a Carbon atom may be chemically more similar to a Nitrogen atom than a Hydrogen atom, which should be reflected in the graph proximity metric. As mentioned in Section~\ref{subsec-notes-web}, there are other definitions of GED that can handle such cases.

% Regarding efficiency of \model, computing GED is a preprocessing step whose time complexity is in Section~\ref{subsec-ged-algo}. The training time complexity of \model is $O(|E|M)$, where $|E|$ is the average number of edges in a training graph pair, and $M$ is the number of training pairs per iteration. Our model is inductive and can be used to embed any unseen graph with time complexity $O(|E|)$.

%% file: sec-other-metric.tex
\section{Impact of Graph Proximity Metrics on \model}

We compare the performance of our framework trained under different graph-graph similarity metrics, with results in Table~\ref{table:metric_result}. It turns out that accuracy scores on GED and MCS are comparable to each other on \imdb, which is reasonable because both GED and MCS are NP-hard metrics that take node and edge correspondence into account, and it has been shown that GED and MCS are equivalent under certain conditions~\cite{bunke1997relation}.

\textsc{Graphlet Kernel} (\textsc{GK})~\cite{shervashidze2009efficient} measures the similarity between two graphs by their shared graphlets, e.g. triangles, rings, etc. When trained using this heuristic-based graph kernel, \model performs worse than when trained using GED and MCS. Note that our model learns from \textsc{GK}, and thus in theory should achieve an accuracy very similar to \textsc{GK}, which is indeed verified by comparing \model-\textsc{GK} with \textsc{GK}.

Note that GED is not a graph kernel, and the high computational cost prevents us from directly using GED to conduct classification for large-scale graph databases. 
% Regarding the training size, indeed more training pairs lead to better performance, when the model can be better fit. Note that according to Fig. 3, when we only use 0.001% of all the training graph pairs, the performance is still better than many baseline methods, and we observe that the performance has moderate gain with more training data.

\begin{table}
% \small
% \vspace*{-4mm}
\begin{tabular}
{ccccc} \hline
\textbf{Method} & \textbf{\imdb} & \textbf{\nci} \\ \hline
\model-GED & 50.06 & 69.17 \\
\model-MCS & 50.01 & 69.85 \\
\model-\textsc{GK} & 43.17 & 60.05 \\
\textsc{GK} & 43.89 & 62.06 \\
\hline
\end{tabular}
\centering
\caption{Graph classification accuracy of \model trained under different graph proximity metrics (GED, MCS, and \textsc{GK}) as well as \textsc{GK} alone which is a graph kernel method.}
\label{table:metric_result}
% \vspace*{-2mm}
\end{table}

%% file: sec-msna.tex
\section{Analysis of The \textbf{Multi-Scale Node Attention (MSNA)} Mechanism}
\label{sec-msna}

Table~\ref{table:msna} shows how much performance gain our proposed \textbf{Multi-Scale Node Attention (MSNA)} mechanism brings to our model, \model. As can be seen in the table, a simple averaging scheme to generate graph-level embeddings cannot yields much worse performance, compared with the other three mechanisms. The supersource approach is not very bad, but still worse than the attention mechanism which brings learnable components into the model. The \textbf{MSNA} mechanism combines the node embeddings from different GCN layers, capturing structural difference at different scales and yielding the best performance.

\begin{table}
% \small
% \vspace*{-4mm}
\begin{tabular}
{cccc} \hline
\textbf{Method} & \textbf{acc} & \textbf{$\tau$} & \textbf{p@10} \\ \hline
\textsc{\model}-\textsc{avg} & 34.73 & 0.243 & 0.647 \\
\textsc{\model}-\textsc{ssrc} & 46.02 & 0.810 & 0.796 \\
\textsc{\model}-\textsc{na} & 49.51 & 0.851 & 0.810 \\
\textsc{\model}-\textsc{msna} & \textbf{50.06} & \textbf{0.853} & \textbf{0.816} \\
\hline
\end{tabular}
\centering
\caption{\textsc{\model}-\textsc{avg} uses the average of node embeddings to generate graph-level embeddings. \textsc{\model}-\textsc{ssrc} uses supersource machanism as described in Section~\ref{sec-related}. \textsc{\model}-\textsc{na} uses only the node embeddings of the last GCN layer to generate graph-level embeddings, but still uses the node attention mechanism. \textsc{\model}-\textsc{msna} is our full model using three layers of GCN to generate graph-level embeddings. The results are on the \imdb dataset.}
\label{table:msna}
% \vspace*{-2mm}
\end{table}

% \subsection{}

Please note that under all the four settings, the node embeddings layers are exactly the same, i.e. three sequential \gin layers. From \textsc{\model}-\textsc{avg}, we can see that the learnable components in the node embedding model are not enough for good graph-level embeddings. 

It is also worth mentioning that the supersource idea works reasonably well, which can be attributed to the fact that the supersource node is connected to every other node in the graph, so that every node passes information to the supersource node, contributing to a relatively informative graph-level embedding. Compared with the averaging scheme, there is additional MLP transformation on the node embedding of the supersource node after the aggregation of other node embeddings, as indicated by the Equation for \gin in the main paper.

We also conduct experiments which demonstrates that the performance is much worse when only single scales are used as shown in Table~\ref{table:msna_scale}. 

\begin{table}
% \small
% \vspace*{-4mm}
\begin{tabular}
{ccccc} \hline
\textbf{scale} & \textbf{\# \gin layers} & \textbf{acc} & \textbf{$\tau$} & \textbf{p@10} \\ \hline
1 & 3 & 47.33 & 0.775 & 0.789 \\
2 & 3 & 47.33 & 0.838 & 0.797 \\
3 & 3 & 49.51 & 0.851 & 0.810 \\
1,2,3 & 3 & 50.06 & 0.853 & 0.816 \\
1,2,3,4 & 4 & 50.01 & 0.851 & 0.810 \\
1,2,3,4,5 & 5 & 50.67 & 0.854 & 0.815 \\
\hline
\end{tabular}
\centering
\caption{``scale'' denotes the indices of the \gin layers that are used for generating graph-level embeddings. ``\# \gin layers'' indicates the total number of \gin layers stacked sequentially. The results are on the \imdb dataset.}
\label{table:msna_scale}
% \vspace*{-2mm}
\end{table}

%% file: sec-vis.tex
\section{Extra Visualizations}

\begin{figure*}
\centering
\includegraphics[scale=0.28]{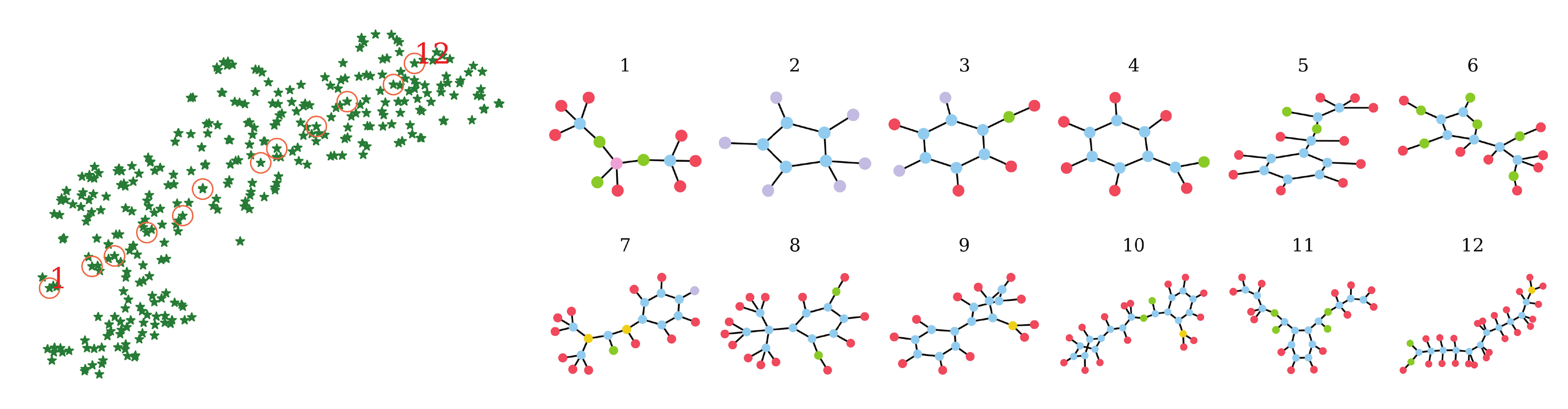}
\caption{Visualization of the \ptc dataset.}
\label{fig:ptc_vis}
\vspace*{-2mm}
\end{figure*}

% \begin{figure*}
% \centering
% \includegraphics[scale=0.3]{imdb_vis}
% \caption{Visualization of the \imdb dataset.}
% \label{fig:imdb_vis}
% \vspace*{-2mm}
% \end{figure*}

\begin{figure*}
\centering
\includegraphics[scale=0.28]{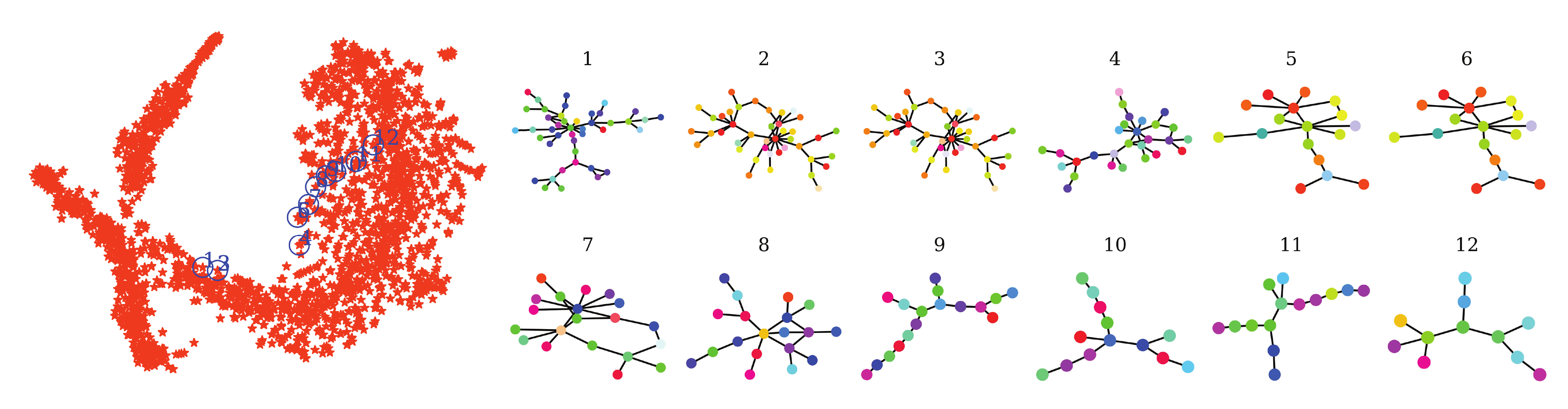}
\caption{Visualization of the \web dataset.}
\label{fig:web_vis}
\vspace*{-2mm}
\end{figure*}

\begin{figure*}
\centering
\includegraphics[scale=0.28]{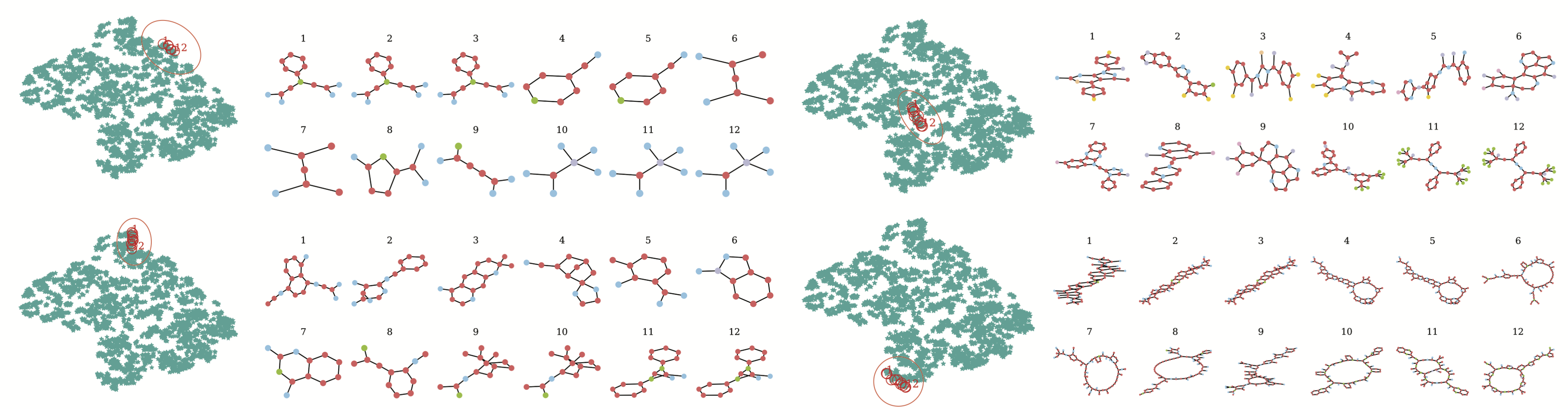}
\caption{Visualization of the \nci dataset.}
\label{fig:nci_vis}
\vspace*{-2mm}
\end{figure*}

\begin{figure*}
\centering
\includegraphics[scale=0.128]{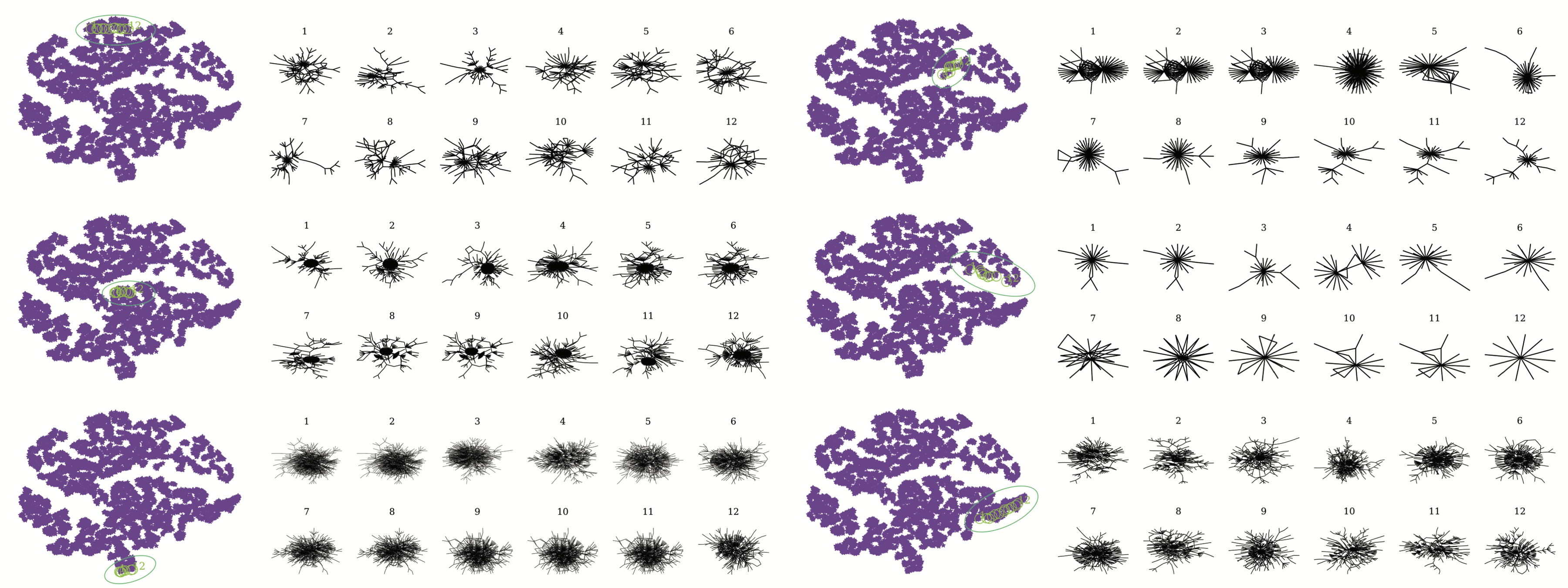}
\caption{Visualization of the \reddit dataset.}
\label{fig:reddit_vis}
\vspace*{-2mm}
\end{figure*}

A few more visualizations are included in Fig.~\ref{fig:ptc_vis}, \ref{fig:imdb_vis}, \ref{fig:web_vis}, \ref{fig:nci_vis}, and \ref{fig:reddit_vis}, for the \ptc, \web, \nci, and \reddit datasets used in the main paper.

\label{sec-conc}